\documentclass[conference]{IEEEtran}
\IEEEoverridecommandlockouts

\usepackage{cite}
\usepackage{amsmath,amssymb,amsfonts}
\usepackage{algorithmic}
\usepackage{graphicx}
\usepackage{textcomp}
\usepackage{xcolor}
\usepackage{booktabs}       
\usepackage{multirow}
\usepackage{algorithm}
\usepackage{algorithmic}
\usepackage{url}
\usepackage{placeins}
\def\BibTeX{{\rm B\kern-.05em{\sc i\kern-.025em b}\kern-.08em
    T\kern-.1667em\lower.7ex\hbox{E}\kern-.125emX}}
\begin{document}

\title{Accelerating Targeted Hard-Label Adversarial Attacks in Low-Query Black-Box Settings
}

\author{
\IEEEauthorblockN{
Arjhun Swaminathan\IEEEauthorrefmark{1}\IEEEauthorrefmark{2}\IEEEauthorrefmark{3}\thanks{*Corresponding author.}
}
\IEEEauthorblockA{University of Tübingen, Germany\\
\texttt{arjhun.swaminathan@uni-tuebingen.de}
}
\and
\IEEEauthorblockN{
Mete Akgün\IEEEauthorrefmark{2}\IEEEauthorrefmark{3}
}
\IEEEauthorblockA{University of Tübingen, Germany\\
\texttt{mete.akguen@uni-tuebingen.de}
}
\thanks{\IEEEauthorrefmark{2}Medical Data Privacy and Privacy-preserving Machine Learning (MDPPML), University of Tübingen, Germany.}
\thanks{\IEEEauthorrefmark{3}Institute for Bioinformatics and Medical Informatics (IBMI), University of Tübingen, Germany.}
}

\maketitle

\begin{abstract}
Deep neural networks for image classification remain vulnerable to adversarial examples --- small, imperceptible perturbations that induce misclassifications. In black-box settings, where only the final prediction is accessible, crafting targeted attacks that aim to misclassify into a specific target class is particularly challenging due to narrow decision regions. Current state-of-the-art methods often exploit the geometric properties of the decision boundary separating a source image and a target image rather than incorporating information from the images themselves. In contrast, we propose Targeted Edge-informed Attack (TEA), a novel attack that utilizes edge information from the target image to carefully perturb it, thereby producing an adversarial image that is closer to the source image while still achieving the desired target classification. Our approach consistently outperforms current state-of-the-art methods across different models in low query settings (nearly 70\% fewer queries are used), a scenario especially relevant in real-world applications with limited queries and black-box access. Furthermore, by efficiently generating a suitable adversarial example, TEA provides an improved target initialization for established geometry-based attacks.
\end{abstract}


\section{Introduction}\label{sec:introduction}
Deep neural networks have achieved remarkable performance in image classification tasks, powering applications from autonomous systems \cite{bojarski2016end,chen2015deepdriving} to medical diagnostics \cite{esteva2017dermatologist,he2016deep}. However, they have repeatedly been shown to be vulnerable to adversarial examples \cite{goodfellow2014explaining,szegedy2013intriguing}. These are small, often imperceptible perturbations to a correctly classified image that cause a misclassification. Although many of these attacks assume white-box access to a model’s internals, hard-label (decision-based) attacks offer a more challenging yet practical setting where only the top-1 predicted label is observed. This limited-feedback scenario commonly arises in commercial APIs \cite{ilyas2018black}. In this realm, targeted attacks, which push the model’s prediction to a specific target class, are inherently more difficult since the decision regions corresponding to the specific target classes are usually narrower and more isolated. 

In black-box settings, targeted hard-label attacks have become an active area of research, with several techniques proposed in the literature \cite{cheng2019sign, cheng2019improving,vo2021ramboattack,chen2017zoo,chengquery}, with state-of-the-art methods relying on the geometry of the decision boundary separating the source image from the target class. Geometry-informed attacks traverse in a lower-dimensional space and fall into two categories: \emph{Boundary Tracing Attacks} and \emph{Gradient Estimation Attacks}. Boundary Tracing Attacks (\cite{brendel2017decision, brunner2019guessing, maho2021surfree,suya2020hybrid}) perform walks along the decision boundary while Gradient Estimation Attacks (\cite{chen2020hopskipjumpattack, li2020qeba, liu2019geometry, ma2021finding, reza2023cgba, tashiro2020diversity, wang2022triangle}) perform the same walks but using information about the approximate tangent/normal to the decision boundary in a local neighborhood to their adversarial image. Although powerful for local refinement, both approaches tend to burn through queries when the source image lies far from a given adversarial image in a target class, wasting many queries before reaching a narrow region where local geometry can more effectively be leveraged. 

Further, in practice, many real-world scenarios such as commercial pay-per-query APIs, impose severe constraints on the number of queries that can be made to a target model in a specified time. Often practical query limits may be on the order of a few hundred to fewer than a few thousand queries. Under these conditions, the limited feedback available makes it difficult to effectively use information about the local decision boundary geometry in the early stages of an attack. When the decision space is still wide, movement along restricted lower dimensional spaces leads to limited incremental progress, and gradient estimation methods often use queries in estimating local gradients by sampling predictions in a local neighborhood instead of moving. This raises a pressing need to develop methods that are efficient in the low-query regime. In a high query setting, this would also lead us to a good starting point to employ the geometry-informed methods since we arrive at a good adversarial point quickly and can use the geometry-informed information more effectively.
To this end, we propose Targeted Edge-Informed Attack (TEA), a novel targeted adversarial attack designed for hard-label black-box scenarios within a restricted query budget. Rather than depending on the local properties of the decision boundary, TEA leverages the intrinsic features of the target image itself - specifically, the edge information obtained using Sobel filters \cite{sobel1978neighborhood} help identify prominent structural features in the target image. Edges encode high‑magnitude spatial gradients that delineate object boundaries \cite{sobel1978neighborhood}. Research has shown that early layers fire on oriented edge filters \cite{bassett2020color,zeiler2014visualizing}, and that shape/edge cues remain predictive even when textures are suppressed \cite{geirhos2018imagenet}. The core idea is to preserve these low-level features, while applying perturbations to the non-edge regions of an image, allowing us to stay in the target class while pushing the adversarial image towards the source image. When our progress plateaus, evidenced by a series of consecutive queries that fail to achieve further reduction in distance while maintaining target class prediction, one can switch to current state-of-the-art geometry-informed methods for a local refinement procedure. Hence we make the following contributions:

\textbf{TEA:} We introduce an edge-informed perturbation strategy, enabling rapid progress toward the source image in the early stages of an adversarial attack when limited queries are available. TEA involves a two-step process: First, a global edge-informed search is performed, and then, edge-informed updates are applied to small patches using Gaussian weights.

\textbf{Empirical validation under strict query budgets:} We perform extensive evaluations on the ImageNet validation dataset \cite{deng2009imagenet} across four architectures (ResNet‑50 \cite{he2016deep}, ResNet‑101 \cite{he2016deep}, VGG16 \cite{simonyan2014very}, and ViT \cite{dosovitskiy2020image}) and an adversarially trained architecture (ResNet-50 \cite{robustness}). Our attack consistently outperforms existing state‑of‑the‑art hard‑label methods, including HSJA, Adaptive History‑driven Attack (AHA) \cite{li2021aha}, CGBA, and CGBA‑H, under realistic query budgets (fewer than 1000 queries). To achieve a 60\% reduction in distance from a target image to a source image, TEA required on average 251 queries across the four models - 70\% fewer than AHA (the second fastest), which required 845 queries.

The rest of this paper is structured as follows. Section 2 reviews related work on targeted hard-label adversarial attacks. Section 3 describes our methodology, while Section 4 presents experimental results. Finally, Section 5 outlines directions for future research.

\section{Related Works}

Early work in the realm of targeted hard-label adversarial attacks consisted of seminal work: Boundary Attack (BA) \cite{brendel2017decision}, which proposed a method of traversing the decision boundary that separates an adversarial image from a source image. Building on this framework, BiasedBA \cite{brunner2019guessing} incorporated directional priors, such as perceptual and low-frequency biases, to restrict the search to more promising regions. BA with Output Diversification Strategy (BAODS) \cite{tashiro2020diversity} integrates diverse gradient-like signals into the exploration process of BA, thereby strengthening the original method. 

Following these foundational methods, Hybrid Attack (HA) \cite{suya2020hybrid} employed a combination of heuristic search strategies to balance global exploration and local refinement. Advancing toward more gradient-centric techniques, qFool \cite{liu2019geometry} leverages the local flatness of decision boundaries to streamline the attack process. Meanwhile, HopSkipJumpAttack (HSJA) \cite{chen2020hopskipjumpattack} locally approximates the normal to the decision boundary to “jump off” from it before progressing toward the source image. Complementarily, Tangent Attack exploits locally estimated tangents of the decision surface to steer the adversarial perturbation toward the source image. Addressing the challenges posed by high-dimensional input spaces, Query Efficient Boundary Attack (QEBA) \cite{li2020qeba} projects gradient estimation into lower-dimensional subspaces, such as the frequency domain, thus significantly reducing the number of required queries.

Incorporating historical query information, Adaptive History-driven Attack (AHA) \cite{li2021aha} adapts its search trajectory based on previous successes and failures, while Decision-based query Efficient Adversarial Attack based on boundary Learning (DEAL) \cite{chen2020hopskipjumpattack} employs an evolutionary strategy that concentrates queries on promising regions of the input space. SurFree \cite{maho2021surfree} demonstrated that using 2D planes and semicircular trajectories toward the source image was an effective strategy. This was based on the fact that under the assumption of a flat decision boundary, the point on the boundary that is closest to the source image is precisely where the semicircular trajectory intersects it. Building on this idea, Curvature-Aware Geometric black-box Attack (CGBA) and its variant that is more suited for targeted attacks - CGBA-H \cite{reza2023cgba}, use normal estimation at the decision boundary to select the traversal plane. To the best of our knowledge, CGBA-H serves as the current state of the art in decision-based targeted adversarial attacks.

\section{Methodology}\label{sec:methodology}
In this section, we formalize the hard-label attack setting and detail our attack, which drastically reduces early query cost, especially when the target image is far from the source image, and exists in a wider decision space. Once the adversarial image achieves a good distance and reaches a narrow decision space, one can use the image as an initialization and continue refining with existing geometric-based methods. 

\textbf{Problem Statement.} We consider a hard-label image classifier modeled by
\begin{equation}
  f: [0,1]^{C \times H \times W} \to \mathbb{R}^K,
\end{equation}
 where $C$ denotes the number of color channels, $H$ and $W$ are the image height and width, and the classifier distinguishes among $K$ classes. For any query image $x$, we do not observe its continuous output (e.g., logits or probabilities) but only the predicted label index
\begin{equation}
  \hat{y}(x) = \arg\max_{1 \leq k \leq K}[f(x)]_k.
\end{equation}

Let $x_s$ be a \emph{source} image correctly classified as $y_s$. In a \emph{targeted} attack, we start with a \emph{target} image $x_t$ (correctly classified as $y_t$). Our goal is to find an adversarial image $x_{\mathrm{adv}}$ that is as close as possible to $x_s$ (in the $\ell_2$ norm) while maintaining the target classification, i.e.,
\begin{equation}
  x_{\mathrm{adv}} = \arg\min_{x} \| x - x_s \|_2, \quad \text{subject to} \quad \hat{y}(x)=y_t.
\end{equation}

We propose a two-part procedure for perturbing the target image $x_t$ to get closer to $x_s$ while maintaining adversarial requirements. First, a \emph{global} edge-informed search coarsely aligns major image regions. Second, a \emph{patch-based} edge-informed search further perturbs local regions in our adversarial image. Note that neither step leverages local decision boundary geometry or gradient estimation, which are typically beneficial only when the adversarial image lies near a narrow decision space and when movements towards the source image are unlikely to maintain classification. This approach avoids burning queries in a wider decision space.

\subsection{Global Edge-Informed Search}
\label{subsec:global}
The global perturbation step aims to coarsely align $x_t$ towards $x_s$ by modifying predominantly smooth, non-edge areas, thereby preserving crucial edge structures that help maintain target classification.

\textbf{Soft Edge Mask.}
To this end, we detect edges in $x_t$ using the Sobel operator \cite{sobel1978neighborhood}, as depicted in Figure \ref{architecture}. A subsequent blurring operation yields a \emph{soft edge mask} $M_{\mathrm{edge}}$, where $M_{\mathrm{edge}}(i,j) \in [0,1]$ has values close to 1 at edge locations and gradually transitions to 0 in smoother regions. We describe this in Algorithm \ref{alg:edge-mask}.

\begin{algorithm}[!ht]
\small
\caption{Soft Edge Mask}
\label{alg:edge-mask}
\begin{algorithmic}[1]
\STATE \textbf{Inputs:} Image $x$, edge thresholds $\{T_\ell, T_h\}$, Gaussian blur kernel size $b$, intensity factor $\gamma$, small constant $\epsilon$\\[1mm]
\STATE \textbf{Output:} Soft edge mask $M_{\mathrm{edge}}$ \\[1mm]
\STATE $x^{\text{gray}} \gets \text{GrayScale}(x)$
\STATE $(s_x, s_y) \gets (\text{Sobel}(x^{\text{gray}}, 0),\; \text{Sobel}(x^{\text{gray}}, 1))$
\STATE $G \gets (s_x^2 + s_y^2)^{1/2}$
\STATE $G \gets 255 \cdot\left(G/(max_{i,j}(G)+\epsilon)\right)$
\STATE $\text{edge\_mask}(i,j) \gets
  \begin{cases}
    255, & \text{if } T_\ell \le G(i,j) \le T_h, \\
    0,   & \text{otherwise}
  \end{cases}$
\STATE $\text{blurred} \gets \text{GaussianBlur}(\text{edge\_mask}, (b,b))$
\STATE $\text{norm} \gets \text{Normalize}(\text{blurred})$
\STATE $M_{\mathrm{edge}} \gets \gamma \cdot \text{norm}$\\[1mm]
\RETURN $M_{\mathrm{edge}}$
\end{algorithmic}
\end{algorithm}

\textbf{Global Interpolation.}
Starting from $x_0 = x_t$, we perform iterative updates
\begin{equation}
  x_{k+1} \leftarrow x_k + \alpha  (x_s - x_k) \odot \left(I - M_{\mathrm{edge}}\right),
\end{equation}
where $\odot$ denotes the element-wise (Hadamard) product. The scaling factor $\alpha$ is optimized via a momentum based search. Masking out the edge regions in this update ensures that the interpolation primarily affects smooth areas, thus preserving the overall structure necessary for target classification. The complete procedure is summarized in Algorithm~\ref{alg:global2}. Once the improvements begin to stagnate, we transition to a local, patch-based refinement.

\begin{algorithm}[!ht]
\small
\caption{Global Edge-Informed Search}
\label{alg:global2}
\begin{algorithmic}[1]
\STATE \textbf{Inputs:} Source image $x_s$, target image $x_t$, soft edge mask $M_{\mathrm{edge}}$, tolerance $\tau$, maximum queries $qc_{\max}$, initial step factor $\eta$, momentum $\mu$\\[1mm]
\STATE \textbf{Output:} Adversarial image $x_{\mathrm{adv}}$ such that $\hat{y}(x_{\mathrm{adv}})=y_t$\\[1mm]
\STATE $x_{\text{current}} \gets x_t$, $v \gets 0$, and $d \gets x_s - x_t$
\STATE Set step size: $s \gets \|d\|_2\cdot \eta$, and initialize query count $qc \gets 0$
\STATE \textbf{while }{$qc < qc_{\max}$ \textbf{and} $\|s\| \ge \tau$}
    \STATE  $\quad$$v \gets \mu\cdot v + (1-\mu) \cdot d$
    \STATE  $\quad$$x_{\text{next}} \gets x_{\text{current}} + s\cdot (v \odot (I-M_{\mathrm{edge}}))$
    \STATE  $\quad$$qc \gets qc + 1$
    \STATE $\quad$ \textbf{if }$\hat{y}(x_{\text{next}})=y_t$
        \STATE  $\quad$ $\quad$$x_{\text{current}} \gets x_{\text{next}}$
        \STATE  $\quad$ $\quad$$s \gets 1.1 \cdot s$ 
    \STATE  $\quad$\textbf{else}
        \STATE  $\quad$ $\quad$\textbf{break}\\[1mm]
\RETURN $x_{\text{current}}$
\end{algorithmic}
\end{algorithm}

\subsection{Patch-Based Edge-Informed Search}
\label{subsec:patch}
\begin{figure*}[t]
\centering
\includegraphics[width=0.7375\textwidth]{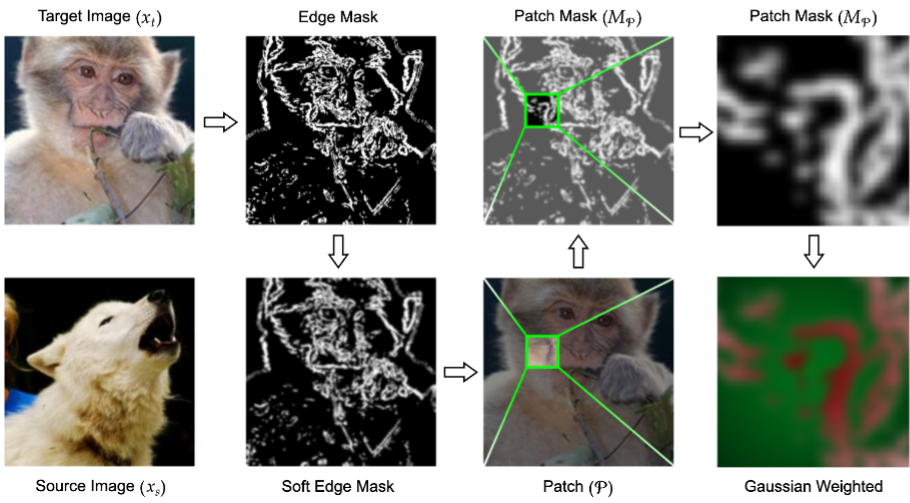}
\caption{\textbf{Overview of Patch-Based Edge-Informed Search.} Edge information from the target image, obtained via the Sobel operator, is first blurred to generate a soft edge mask. A square patch is then selected and a Gaussian weighting function is applied. In the bottom right panel, the intensity of the modification is illustrated: dark red regions remain largely unchanged, while light green regions receive a more pronounced update. The lack of changes near the patch borders helps prevent the introduction of artificial edges.}
\label{architecture}
\end{figure*}

After the global interpolation, some subregions of the image may still display significant discrepancies from $x_s$. In the patch-based refinement, we then partition the image into randomly selected patches of random sizes
\begin{equation}
\mathcal{P} \subseteq \{1,\dots,H\} \times \{1,\dots,W\},
\end{equation}
and for each patch, construct a local soft edge mask $M_{\mathcal{P}}$ by restricting $M_{\mathrm{edge}}$ to $\mathcal{P}$. Next, starting from our adversarial image $x_k$, we apply a local interpolation
\begin{equation}
  \widetilde{x}(\beta) = x_k + \beta (x_s - x_k) \odot G \odot (I-M_{\mathcal{P}}),
\end{equation}
and search for the largest $\beta$ such that the classifier still predicts the target label, $\hat{y}(\widetilde{x}(\beta)) = y_t$. Here, $G$ is a Gaussian weighting function over the patch that smoothly downweights updates near patch borders, helping to avoid artificial edges introduced by patch boundaries. We repeat this for different patches until a termination criterion is met (e.g., 25 consecutive iterations with no further improvement). The corresponding illustration is depicted in Figure \ref{architecture} and the relevant pseudocode is provided in Algorithm~\ref{alg:patch}.

\begin{figure*}[!ht]
\centering
\includegraphics[width=0.75\textwidth]{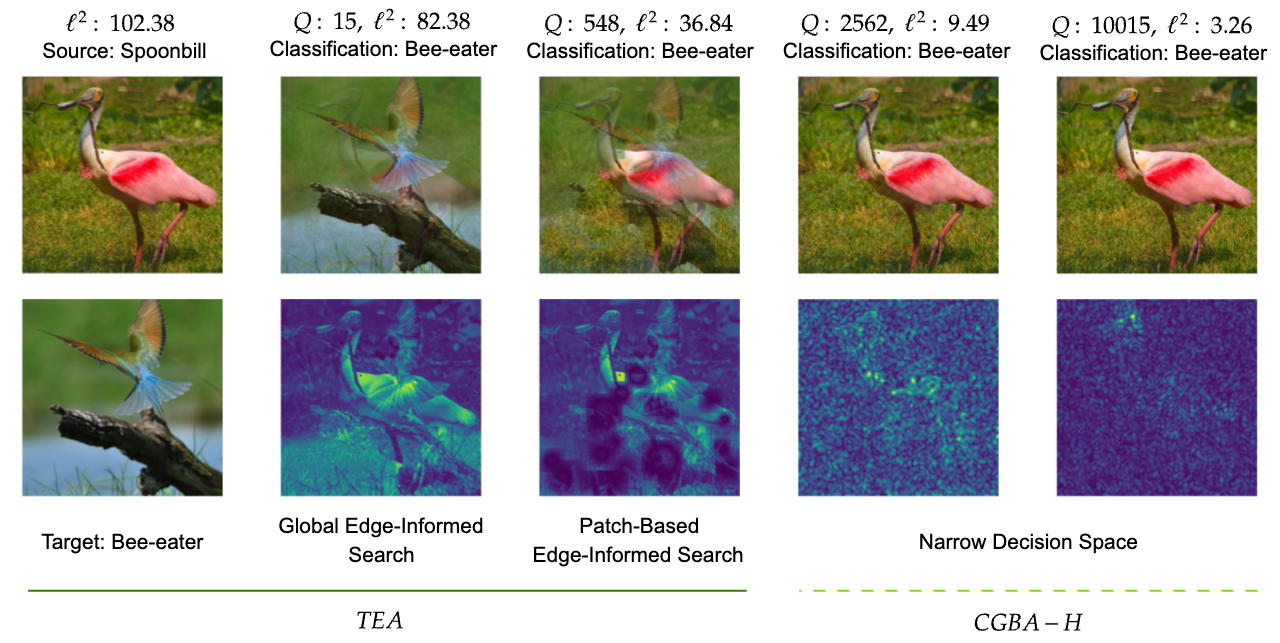}
\caption{\textbf{Visualization of TEA on a source–target image pair.} The target image (initially classified as \emph{Bee-eater}) is perturbed to resemble the source image (classified as \emph{Spoonbill}), while preserving its original \emph{Bee-eater} label. Global Edge-Informed Search efficiently applies edge-aware perturbations using only $15$ queries to achieve a $\approx$20\% reduction in distance to the source. Patch-Based Edge-Informed Search introduces localized, edge-aware modifications to small image regions, as seen in the hotspot of changes. Further refinement utilizing CGBA-H is illustrated in the narrow decision space.}
  \label{fig:TEA}
\end{figure*}

Figure \ref{fig:TEA} offers a visual overview of TEA. The figure depicts the adversarial image throughout our method, and displays a hotspot of individual pixel differences from the source image. 

\begin{algorithm}[!ht]
\caption{Patch-Based Edge-Informed Search}
\label{alg:patch}
\small
\begin{algorithmic}[1]
\STATE \textbf{Inputs:} Source image $x_s$, current adversarial image $x_{\mathrm{adv}}$, soft edge mask $M_{\mathrm{edge}}$ (from \texttt{create\_soft\_edge\_mask}), minimum patch size $p_{\min}$, maximum patch size $p_{\max}$, step factor $\eta$, momentum $\mu$, maximum calls $N_{\max}$
\STATE \textbf{Output:} Refined adversarial image $x_{\mathrm{adv}}$
\STATE $n_{\text{break}} \gets 0$
\STATE \textbf{while } $n_{\text{break}} < 25$
\STATE \quad $D \gets \text{AvgPool}\left(|x_s - x_{\mathrm{adv}}|\right)$ 
\STATE \quad Select high-difference indices from $D$ and choose a random center $(i_c,j_c)$
\STATE \quad $p \gets \mathrm{randInt}(p_{\min},\,p_{\max})$;\quad
  $\mathcal{P} \gets \{(i,j)\mid |i - i_c|\le \lfloor p/2\rfloor,\;|j - j_c|\le \lfloor p/2\rfloor\}$
\STATE \quad $M_{\mathcal{P}}(i,j) \gets 
  \begin{cases}
    1, & (i,j)\in\mathcal{P},\\
    0, & \text{otherwise.}
  \end{cases}$
\STATE \quad $x_{\text{patch}} \gets \mathrm{Patch}(x_{\mathrm{adv}}, \mathcal{P})$
\STATE \quad $m_{\text{patch}} \gets 0,\quad d_{\text{base}} \gets \|x_s - x_{\mathrm{adv}}\|_2$
\STATE \quad \textbf{for } {iteration $=1$ to $N_{\max}$}
\STATE \quad \quad $d_{\text{local}} \gets \mathrm{Patch}(x_s,\mathcal{P}) - x_{\text{patch}}$
\STATE \quad \quad $m_{\text{patch}} \gets \mu \cdot m_{\text{patch}} + (1-\mu) \cdot d_{\text{local}}$
\STATE \quad \quad $s_{\text{patch}} \gets \eta \cdot \|x_s - x_{\mathrm{adv}}\|_2 $
\STATE \quad \quad $G \gets \mathrm{GaussianWeight}((i_c,j_c),\sigma=p/3)$
\STATE \quad \quad $\Delta x \gets s_{\text{patch}} \cdot m_{\text{patch}} \odot \left(G \odot \left(1 - (M_{\mathrm{edge}} \odot M_{\mathcal{P}})\right)\right)$
\STATE \quad \quad $x_{\text{patch}}^{+} \gets x_{\text{patch}} + \Delta x$
\STATE \quad \quad $x_{\mathrm{temp}} \gets x_{\mathrm{adv}}- x_{\mathrm{adv}}\odot M_{\mathcal{P}} + x_{\text{patch}}^{+}\odot M_{\mathcal{P}}$
\STATE \quad \quad $d_{\text{new}} \gets \|x_s - x_{\mathrm{temp}}\|_2$
\STATE \quad \quad \textbf{if }{$d_{\text{new}} \ge 0.999\cdot d_{\text{base}}$}
\STATE \quad \quad \quad \textbf{break} 
\STATE \quad \quad \textbf{if }$\hat{y}(x_{\mathrm{temp}})=y_t$
\STATE \quad \quad \quad $x_{\text{patch}} \gets x_{\text{patch}}^{+}$
\STATE \quad \quad \quad $x_{\mathrm{adv}} \gets \mathrm{Replace}(x_{\mathrm{adv}}, \mathcal{P}, x_{\text{patch}})$
\STATE \quad \quad \quad $d_{\text{base}} \gets d_{\text{new}}$
\STATE \quad \quad \quad $n_{\text{break}} \gets 0$
\STATE \quad \quad \textbf{else }
\STATE \quad \quad \quad $n_{\text{break}} \gets n_{\text{break}} +1$
\STATE \quad \quad \quad \textbf{break} 
\STATE \textbf{return} $x_{\mathrm{adv}}$
\end{algorithmic}
\end{algorithm}

\section{Experiments}
\label{sec:experiments}

In this section, we present our empirical evaluation benchmarking \emph{TEA} against existing targeted hard-label attacks. In what follows, we describe our experimental setup and metrics, then detail the performance of each method under a range of query budgets. Our findings indicate that TEA achieves efficient distance reduction to the source image, particularly in the early query regime, before switching to the current state-of-the-art geometry-based attack CGBA-H for further refinement in a narrow decision space.

\subsection{Setup and Metrics}

\textbf{Computational Resources.} The experiments were executed on a High-performance computing (HPC) Cluster. Each node on the cluster consisted of four NVIDIA GeForce GTX 1080 Ti GPUs (one GPU was allocated per source-target pair for a given attack). 

\textbf{Dataset and Image Pairs.}
We randomly sample 1000 source-target image pairs from the ImageNet ILSVRC2012 validation set, ensuring each pair contains images from distinct classes. All images are resized to $3 \times 224 \times 224$. Each pair $(x_s, x_t)$ contains a source image $x_s$, correctly classified under its label, and a target image $x_t$, also correctly classified under a different label. The goal is to modify $x_t$ to approach $x_s$ under the $\ell_2$ norm while keeping the prediction unchanged.

\textbf{Target Models.}
We evaluate our approach on four well-known classifiers: ResNet50 and ResNet101 (CNNs of varying depth), VGG16 (a CNN composed of repeated convolutional blocks), and ViT (a vision transformer that processes images as sequences of patches). These models represent a diverse set of architectures.

\textbf{Compared Methods.}
We compare our method to four targeted hard-label attacks: HSJA, AHA, CGBA, and CGBA-H.For each method, the $\ell_2$ distance to the source image is recorded after each set of queries, and along with the classification label of the perturbed target image.

\textbf{Evaluation Metrics.} Performance is quantified using three metrics. First, we compute the $\ell_2$ distance from $x_s$ to the adversarial example generated for each of the 1000 image pairs as queries progress. Second, the attack success rate (ASR) is defined as the fraction of image pairs for which an $\alpha\%$ reduction in $\ell_2$ distance between adversarial image and source image is achieved as compared to the target image and source image. Third, we also measure the performance of each method when there is a set fixed low-query budget by measuring the ASRs for all $\alpha$ when each method has used up $500$ queries. Finally, we integrate the $\ell_2$ distance versus queries curve to obtain the area under the curve (AUC), which provides an aggregated measure of how rapidly the distance is reduced.

\textbf{Implementation Details.} Our implementation employs a two-stage process. In the first stage, we perform our proposed methodology TEA, where edge-aware distortions are applied to the target image while preserving its target classification, allowing a rapid reduction in distance to the source image. In the second stage, once the perturbations have brought the image into a narrow decision space, the method switches to CGBA-H for further refinement  (denoted as TEA$^{\ast}$ throughout the study). The last query performed with TEA is referred to as the \emph{turning point} throughout the study. We switch to CGBA-H since it consistently performs better than other methods in a high query setting across different architectures. 

\subsection{Results}
\label{sec:results_imagenet}

\begin{table*}[t]
\caption{Median $\ell_2$ distances computed until the turning point across different architectures. \\Lower $\ell_2$ values denote faster adversarial example generation.}
\label{tab:median_grouped}
\centering
    \begin{tabular}{c c c c c c c c c c c c}
      \toprule
      \multirow{2}{*}{Query} 
        & \multicolumn{5}{c}{ResNet50} 
        & \multirow{2}{*}{Query} 
        & \multicolumn{5}{c}{ResNet101} \\
      \cmidrule(lr){2-6} \cmidrule(lr){8-12}
        & HSJA   & CGBA   & CGBA-H & AHA    & TEA
        &       & HSJA   & CGBA   & CGBA-H & AHA    & TEA \\
      \midrule
100   & 92.797 & 88.857 & 83.104 & 80.438 & \textbf{75.2155} & 100  & 90.962 &  85.623 &81.376 & 82.574 & \textbf{74.5036} \\
200   & 88.715 & 87.165 & 79.348 & 75.344 & \textbf{62.0338} & 200  & 88.194 &  84.246 &74.738 & 77.869 & \textbf{61.3794} \\
300   & 88.604 &  86.236 & 74.168 & 71.054 & \textbf{55.5589} & 300  & 87.954 &  83.750 & 71.874 &70.676 & \textbf{54.9321} \\
400   & 88.199 &  85.579 & 71.479 & 67.091 & \textbf{51.9530} & 400  & 87.356 &  82.483 & 68.803 &67.051 & \textbf{51.3780} \\
500   & 87.724 &  84.897 & 68.671 & 64.172 &\textbf{49.8352} & 500  & 87.222 & 81.543 & 66.169 &  63.717 & \textbf{49.3863} \\
602  & 87.203   & 83.937  & 66.774 & 61.168  & \textbf{45.6000} & 592  & 87.133   & 80.339  & 64.528  & 61.134 & \textbf{45.1390} \\
\midrule
      \multirow{2}{*}{Query} 
        & \multicolumn{5}{c}{VGG16} 
        & \multirow{2}{*}{Query} 
        & \multicolumn{5}{c}{ViT} \\
      \cmidrule(lr){2-6} \cmidrule(lr){8-12}
        & HSJA    & CGBA   & CGBA-H & AHA    & TEA
        &       & HSJA   & CGBA   & CGBA-H & AHA    & TEA \\
      \midrule
100   & 93.966  & 92.154 & 85.805 & 84.770 & \textbf{77.2608} & 100  & 80.454 & 76.885 & 73.074 & 68.863 & \textbf{67.5654} \\
200   & 89.884  & 90.899 & 81.599 & 78.445 & \textbf{63.6530} & 200  & 79.981 & 74.702 & 69.511 &65.501  & \textbf{56.1255} \\
300   & 89.665  & 90.516 & 74.359 & 73.174 & \textbf{56.9954} & 300  & 79.606 &  74.139 & 63.655 & 63.951 &\textbf{50.8059} \\
400   & 88.943  & 89.880 & 70.559 & 68.538 & \textbf{53.4018} & 400  & 79.599 & 72.839 & 61.300 & 59.857 & \textbf{47.8666} \\
500   & 88.919  &  89.563 & 67.745 & 65.569 &\textbf{51.3763} & 500  & 78.761 &  71.291 & 58.598 & 56.572 &\textbf{46.2528} \\
588 & 88.389  & 88.362  & 66.578  & 62.126 & \textbf{47.4840} & 554 & 78.626  & 69.954  & 57.710 & 54.494	& \textbf{43.4220} \\
\bottomrule
\end{tabular}

\end{table*}

\begin{figure*}[t]
    \centering
    \begin{minipage}[b]{0.24\textwidth}
        \centering
        \includegraphics[width=\textwidth]{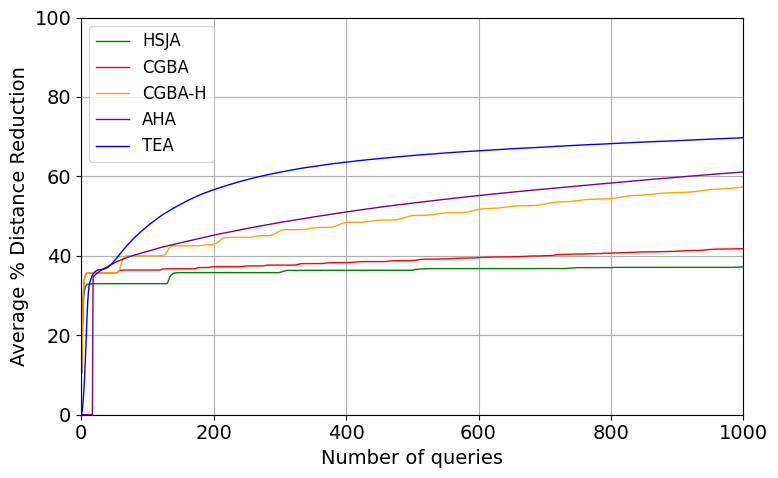}
        \par\small ResNet50
    \end{minipage}\hfill
    \begin{minipage}[b]{0.24\textwidth}
        \centering
        \includegraphics[width=\textwidth]{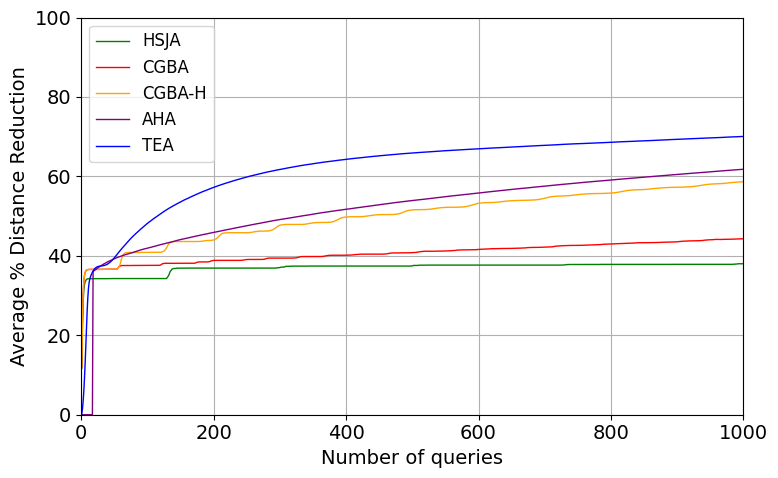}
        \par\small ResNet101
    \end{minipage}\hfill
    \begin{minipage}[b]{0.24\textwidth}
        \centering
        \includegraphics[width=\textwidth]{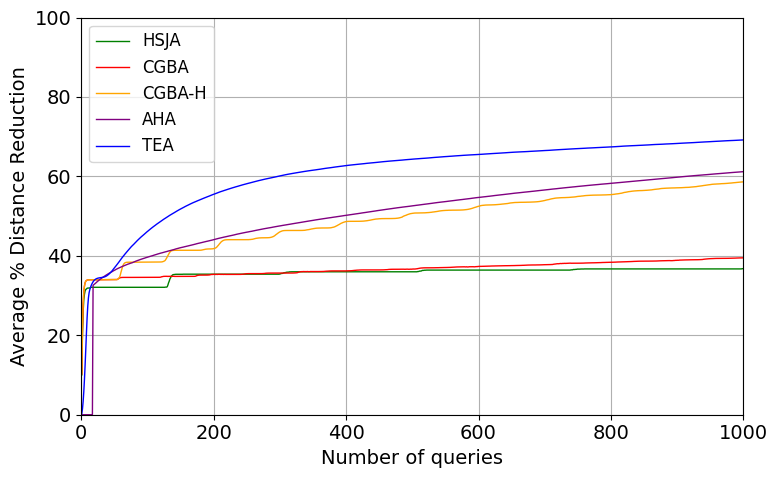}
        \par\small VGG16
    \end{minipage}\hfill
    \begin{minipage}[b]{0.24\textwidth}
        \centering
        \includegraphics[width=\textwidth]{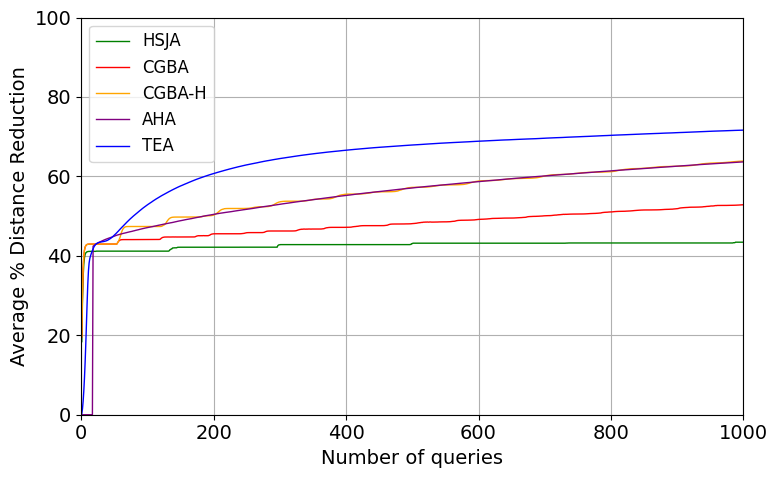}
        \par\small ViT
    \end{minipage}
    \caption{Average $\ell_2$ distance reduction across different architectures in a low-query regime. Higher values indicate improved performance.}
    \label{fig:combined_graphs}
\end{figure*}

\textbf{Average $\ell_2$-Distance vs.\ Queries.}
Table \ref{tab:median_grouped} presents a comparative analysis of the median $\ell_2$ distances achieved in the low-query regime—specifically, within the range defined by the average turning point computed over one thousand image pairs. This analysis underscores the rapid decrease in perturbation norm facilitated by TEA during the early query stages. Figure~\ref{fig:combined_graphs} extends this evaluation by depicting the average percentage reduction in $\ell_2$ distance against queries used. In the plots, the initial inflection point marks TEA’s transition from global exploration to patch-based refinement.

\begin{figure*}[t]
    \centering
    \begin{minipage}[b]{0.24\textwidth}
        \centering
        \includegraphics[width=\textwidth]{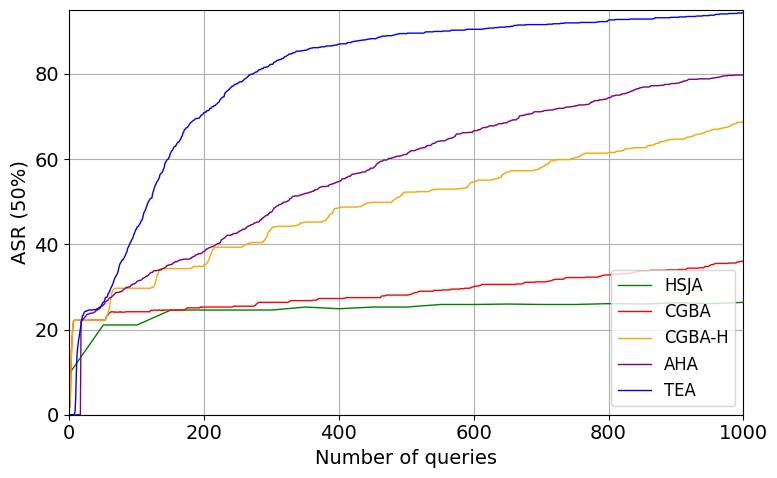}
        \par\small ResNet50
    \end{minipage}\hfill
    \begin{minipage}[b]{0.24\textwidth}
        \centering
        \includegraphics[width=\textwidth]{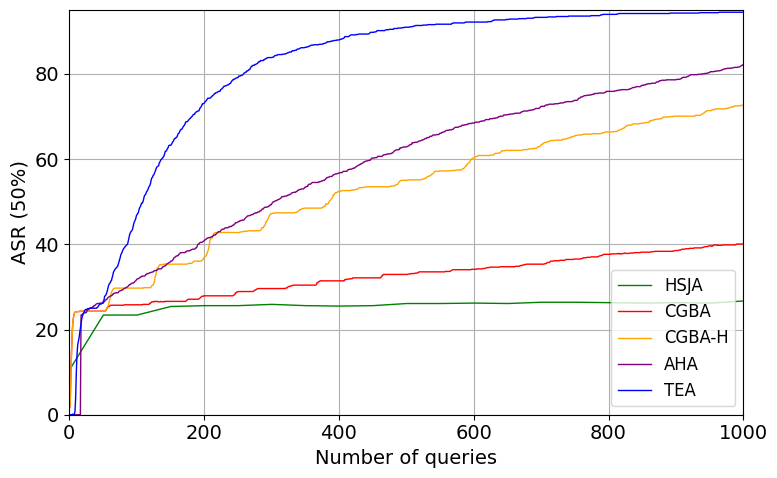}
        \par\small ResNet101
    \end{minipage}\hfill
    \begin{minipage}[b]{0.24\textwidth}
        \centering
        \includegraphics[width=\textwidth]{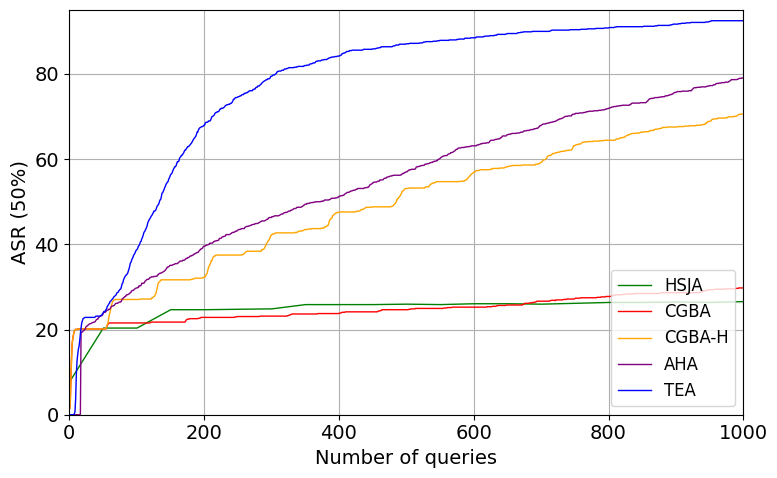}
        \par\small VGG16
    \end{minipage}\hfill
    \begin{minipage}[b]{0.24\textwidth}
        \centering
        \includegraphics[width=\textwidth]{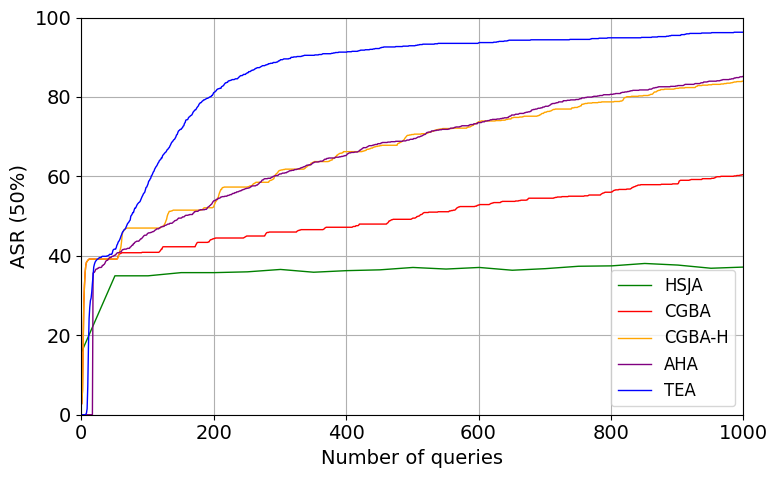}
        \par\small ViT
    \end{minipage}
    \caption{Comparison of ASR of 50\% distance reduction. Higher values indicate that a higher proportion of images reach a distance reduction of 50\% sooner.}
    \label{fig:imagenet_50pct} 
\end{figure*}

\begin{figure*}[t]
    \centering
    \begin{minipage}[b]{0.24\textwidth}
        \centering
        \includegraphics[width=\textwidth]{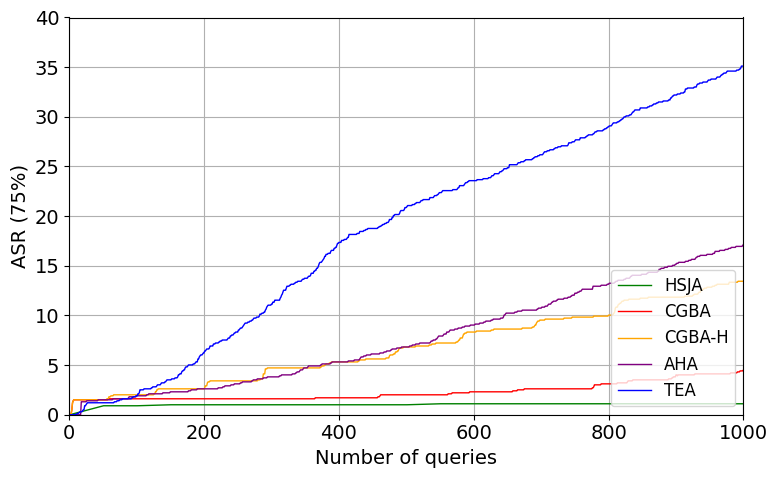}
        \par\small ResNet50
    \end{minipage}\hfill
    \begin{minipage}[b]{0.24\textwidth}
        \centering
        \includegraphics[width=\textwidth]{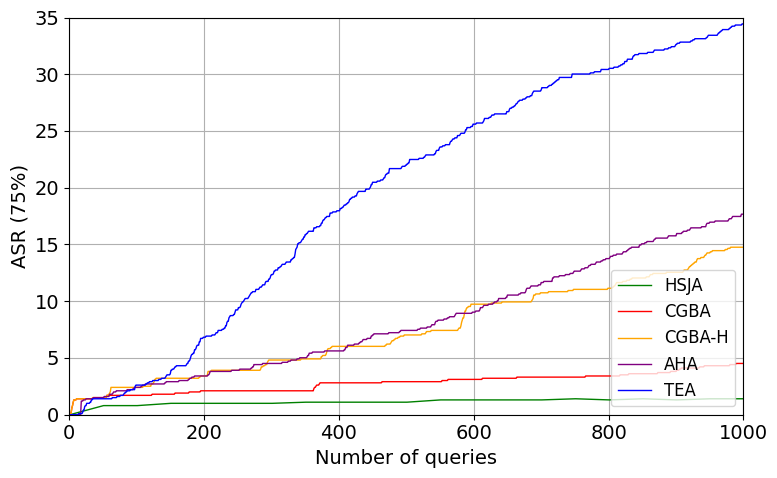}
        \par\small ResNet101
    \end{minipage}\hfill
    \begin{minipage}[b]{0.24\textwidth}
        \centering
        \includegraphics[width=\textwidth]{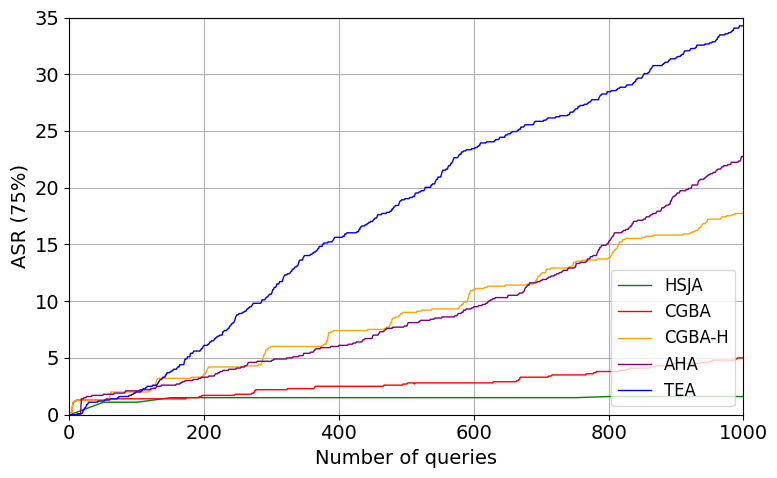}
        \par\small VGG16
    \end{minipage}\hfill
    \begin{minipage}[b]{0.24\textwidth}
        \centering
        \includegraphics[width=\textwidth]{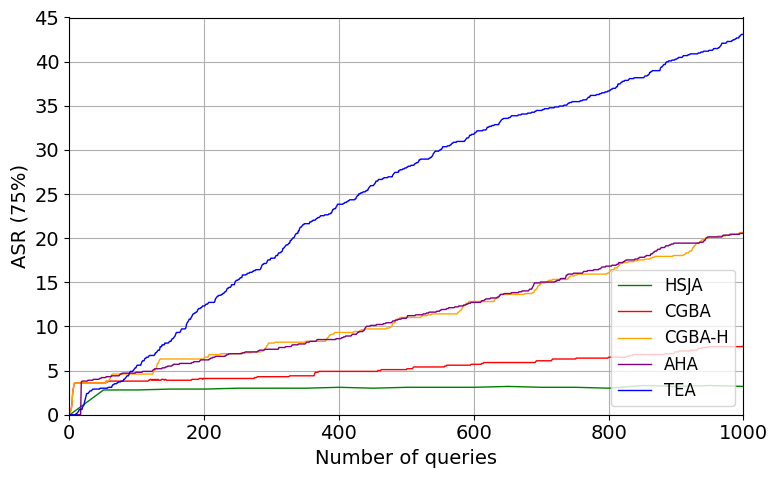}
        \par\small ViT
    \end{minipage}
    \caption{Comparison of ASR of 75\% distance reduction.  Higher values indicate that a higher proportion of images reach a distance reduction of 75\% sooner.}
    \label{fig:imagenet_75pct} 
\end{figure*}

\textbf{Attack Success Rate.}
In Figures~\ref{fig:imagenet_50pct} and ~\ref{fig:imagenet_75pct}, we show how many images reach at least 50\% and 75\% distance reduction over queries respectively. A sharper increase in this fraction signifies that more pairs experience substantial improvement more quickly. Again, once the turning point is reached between a pair, CGBA-H is implemented for further refinement.

\begin{figure*}[t]
    \centering
    \begin{minipage}[b]{0.24\textwidth}
        \centering
        \includegraphics[width=\textwidth]{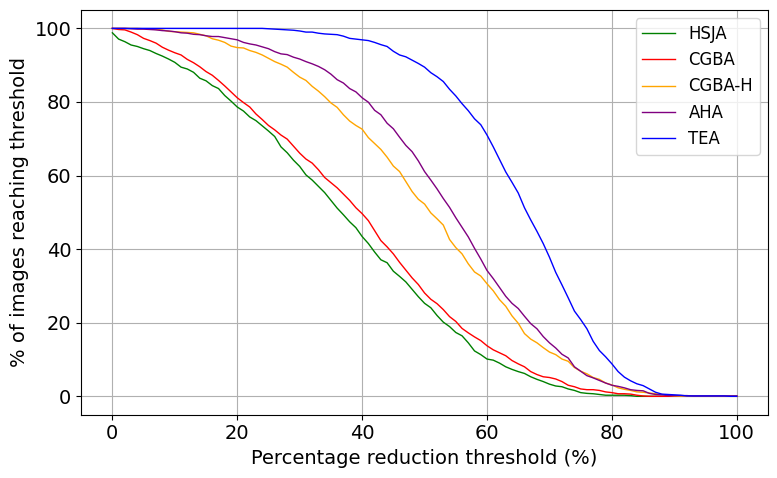}
        \par\small ResNet50
    \end{minipage}\hfill
    \begin{minipage}[b]{0.24\textwidth}
        \centering
        \includegraphics[width=\textwidth]{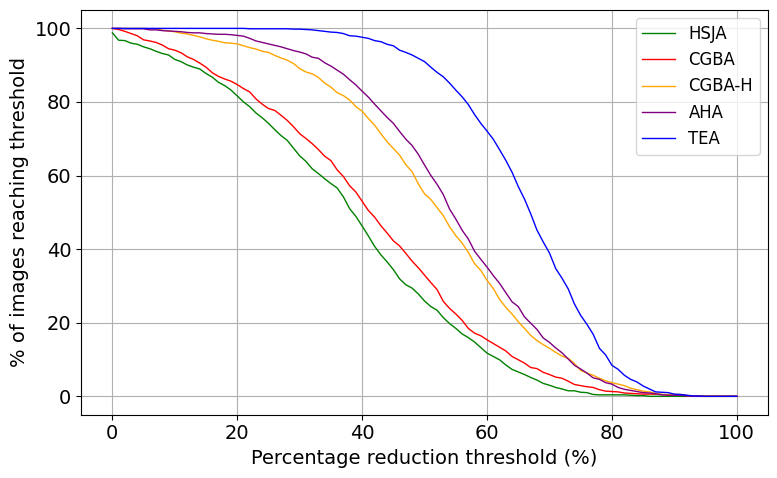}
        \par\small ResNet101
    \end{minipage}\hfill
    \begin{minipage}[b]{0.24\textwidth}
        \centering
        \includegraphics[width=\textwidth]{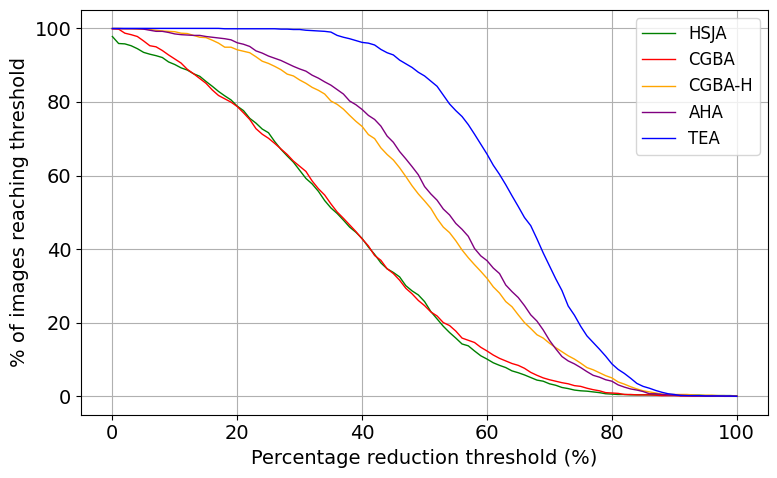}
        \par\small VGG16
    \end{minipage}\hfill
    \begin{minipage}[b]{0.24\textwidth}
        \centering
        \includegraphics[width=\textwidth]{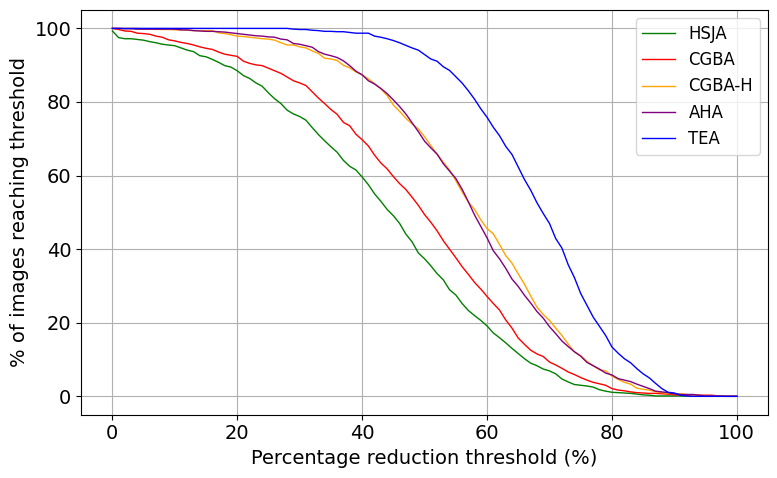}
        \par\small ViT
    \end{minipage}
\caption{Cumulative distribution functions (CDFs) of distance reduction at 500 queries. Each curve represents the fraction of image pairs that achieve a given percentage reduction in the $\ell_2$ distance, with higher values indicating a more effective reduction method.}
    \label{fig:imagenet_threshold_cdf2} 
\end{figure*}

\textbf{Success at a Fixed Low-Query Budget.} Figure \ref{fig:imagenet_threshold_cdf2} reports, at a low-query budget of 500 queries, the proportion of image pairs that reach or exceed various distance-reduction thresholds. TEA maintains a consistently higher proportion of successful pairs across nearly all thresholds, suggesting that its early-stage perturbations secure significant distance reductions more rapidly.

\textbf{AUC Comparisons.}
We assess the efficiency of our aproach by computing the area under the median $\ell_2$-distance vs queries curve (AUC). Table~\ref{tab2:imagenet_auc} presents the AUC values for the low-query regime, up to the turning point. Across all architectures, TEA consistently yields lower AUC values.

\begin{table}[!ht]
\scriptsize
\centering
\caption{Average AUC values computed until the turning point across different architectures. Lower AUC values denote more effective early-stage distance reduction.}
\label{tab2:imagenet_auc}
\begin{tabular}{lccccc}
\toprule
Model      & HSJA        & CGBA        & CGBA-H      & AHA        & TEA \\
\midrule
ResNet50   & 53575.41   & 52130.74   & 45745.09   & 44332.55  & \textbf{35133.14} \\
ResNet101  & 52269.70   & 49487.56   & 43865.80   & 43542.48  & \textbf{34189.94} \\
VGG16      & 53053.06   & 53226.27   & 45261.02   & 44756.50  & \textbf{35790.23} \\
ViT        & 44377.75   & 41097.13   & 36788.92   & 37356.25  & \textbf{30337.35} \\
\bottomrule
\end{tabular}
\end{table}

\begin{figure}[t]
\centering
\resizebox{\columnwidth}{!}{%
\begin{tabular}{@{}cc@{}}
\includegraphics[width=.48\linewidth]{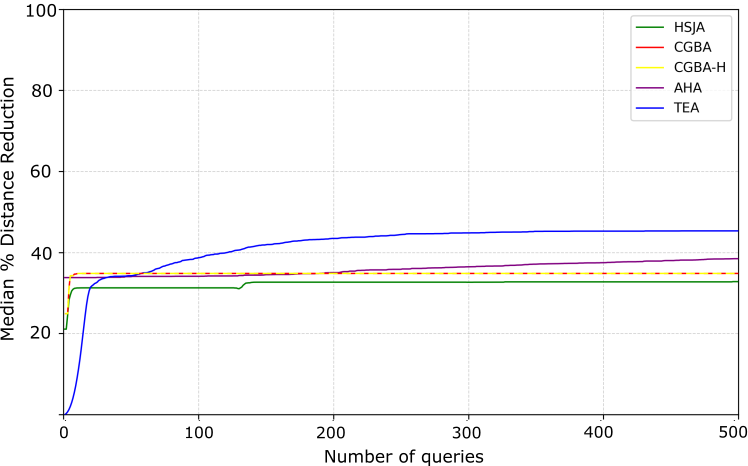} &
\includegraphics[width=.48\linewidth]{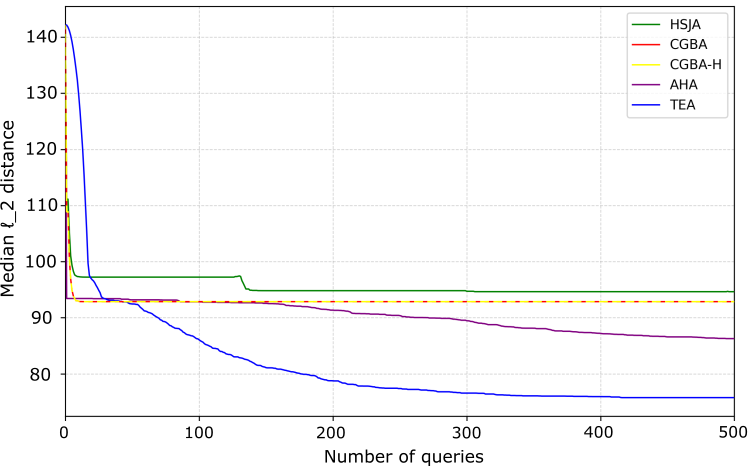}
\end{tabular}%
}
\caption{Comparison on an adversarially trained model. Left: median percentage decrease in $\ell_2$ distance against number of queries, with higher values indicating a more effective reduction method. Right: median $\ell_2$ distance against number of queries, with lower values indicating a more effective reduction method.}
\label{fig:adv_trained_combined}
\end{figure}

\textbf{Evaluation on an Adversarially Trained Model.} In Figure \ref{fig:adv_trained_combined}, we present a comparative analyses of performance against an adversarial trained model (Resnet50 from MardyLab \cite{robustness}) in the low-query region. We see that TEA consistently achieves more distortion, while also noting that all methods perform significantly worse compared to the standard Resnet50 model as seen in Figure \ref{fig:combined_graphs}.

\textbf{Edge Ablation Study} To better understand the role of edge preservation in our method, we perform an ablation study by comparing three variants of TEA under identical settings. TEA perturbs non-edge pixels, preserving edge regions with a soft edge mask. INV-TEA inverts TEA’s soft edge mask to perturb primarily edge-like pixels. HALF-TEA perturbs non-edge and edge regions equally, using the same editable-pixel budget as TEA, and serves as a simple control baseline. Table~\ref{tab:edge_ablation} summarizes results in the low-query regime: median $\ell_2$ distance, ASR at $50\%$ distance reduction, and ASR at $75\%$ distance reduction. Across all four architectures, TEA achieves the lowest median $\ell_2$ and the highest success rates overall; HALF-TEA is consistently in between, and INV-TEA performs the worst, indicating that prioritizing non-edge regions proves crucial for early attack progress while preserving the semantic structure.

\begin{table*}[t]
\caption{Edge ablation in the low-query regime. \\Per architecture and budget we report median $\ell_2$ (lower is better), ASR at 50\%, and ASR at 75\% (higher is better).}
\label{tab:edge_ablation}
\centering
\scriptsize
\resizebox{\textwidth}{!}{%
\begin{tabular}{c c c c c c c c c c c c c c c c c c c c}
\toprule
\multirow{3}{*}{Query}
& \multicolumn{9}{c}{ResNet-50}
& \multirow{3}{*}{Query}
& \multicolumn{9}{c}{ResNet-101} \\
\cmidrule(lr){2-10}\cmidrule(lr){12-20}
& \multicolumn{3}{c}{$\ell_2$} & \multicolumn{3}{c}{ASR@50\%} & \multicolumn{3}{c}{ASR@75\%}
&
& \multicolumn{3}{c}{$\ell_2$} & \multicolumn{3}{c}{ASR@50\%} & \multicolumn{3}{c}{ASR@75\%} \\
\cmidrule(lr){2-4}\cmidrule(lr){5-7}\cmidrule(lr){8-10}\cmidrule(lr){12-14}\cmidrule(lr){15-17}\cmidrule(lr){18-20}
& TEA & Half & Inv & TEA & Half & Inv & TEA & Half & Inv
&
& TEA & Half & Inv & TEA & Half & Inv & TEA & Half & Inv \\
\midrule
100 & \textbf{75.1377} & 87.3465 & 87.5367 & \textbf{44.0} & 23.6 & 22.8 & \textbf{2.0} & 0.2 & 1.3
    & 100 & \textbf{74.4637} & 86.1526 & 86.8610 & \textbf{44.4} & 24.1 & 23.2 & \textbf{2.8} & 0.3 & 1.0 \\
200 & \textbf{61.8307} & 73.1697 & 73.3559 & \textbf{70.5} & 49.7 & 45.2 & \textbf{6.5} & 0.7 & 2.6
    & 200 & \textbf{61.3517} & 71.7322 & 72.6071 & \textbf{72.4} & 53.3 & 47.3 & \textbf{6.9} & 0.9 & 3.1 \\
300 & \textbf{55.3574} & 65.0474 & 66.5448 & \textbf{81.8} & 67.8 & 61.8 & \textbf{10.7} & 3.3 & 4.4
    & 300 & \textbf{54.9709} & 63.5007 & 65.8087 & \textbf{83.9} & 69.7 & 62.0 & \textbf{11.8} & 3.2 & 4.7 \\
400 & \textbf{51.8838} & 60.0257 & 63.2027 & \textbf{86.0} & 77.6 & 69.0 & \textbf{15.6} & 5.9 & 6.1
    & 400 & \textbf{51.3084} & 58.5084 & 62.4517 & \textbf{87.5} & 77.6 & 69.0 & \textbf{17.1} & 6.7 & 6.0 \\
500 & \textbf{49.8877} & 56.8181 & 61.6506 & \textbf{87.5} & 81.8 & 71.3 & \textbf{19.2} & 9.5 & 6.9
    & 500 & \textbf{49.3153} & 55.3806 & 60.8018 & \textbf{89.6} & 82.6 & 72.2 & \textbf{19.7} & 9.5 & 6.5 \\
\midrule
\multirow{3}{*}{Query}
& \multicolumn{9}{c}{VGG16}
& \multirow{3}{*}{Query}
& \multicolumn{9}{c}{ViT} \\
\cmidrule(lr){2-10}\cmidrule(lr){12-20}
& \multicolumn{3}{c}{$\ell_2$} & \multicolumn{3}{c}{ASR@50\%} & \multicolumn{3}{c}{ASR@75\%}
&
& \multicolumn{3}{c}{$\ell_2$} & \multicolumn{3}{c}{ASR@50\%} & \multicolumn{3}{c}{ASR@75\%} \\
\cmidrule(lr){2-4}\cmidrule(lr){5-7}\cmidrule(lr){8-10}\cmidrule(lr){12-14}\cmidrule(lr){15-17}\cmidrule(lr){18-20}
& TEA & Half & Inv & TEA & Half & Inv & TEA & Half & Inv
&
& TEA & Half & Inv & TEA & Half & Inv & TEA & Half & Inv \\
\midrule
100 & \textbf{77.4206} & 90.7514 & 89.8802 & \textbf{38.1} & 17.6 & 20.8 & \textbf{2.0} & 0.3 & 0.8
    & 100 & \textbf{67.4874} & 76.4322 & 77.6473 & \textbf{57.8} & 43.4 & 40.4 & \textbf{5.2} & 0.9 & 2.8 \\
200 & \textbf{63.7679} & 75.3909 & 75.5661 & \textbf{66.3} & 45.0 & 40.8 & \textbf{5.8} & 0.7 & 2.3
    & 200 & \textbf{56.2801} & 63.8529 & 66.0835 & \textbf{80.0} & 69.8 & 60.8 & \textbf{11.8} & 3.6 & 6.7 \\
300 & \textbf{57.0986} & 66.5727 & 68.5222 & \textbf{78.7} & 64.8 & 56.9 & \textbf{10.8} & 2.9 & 3.9
    & 300 & \textbf{50.8057} & 57.2091 & 60.8559 & \textbf{88.1} & 82.8 & 71.0 & \textbf{17.5} & 7.7 & 9.7 \\
400 & \textbf{53.4352} & 61.0717 & 65.1434 & \textbf{83.3} & 74.9 & 62.9 & \textbf{16.8} & 7.4 & 5.0
    & 400 & \textbf{47.9956} & 53.3073 & 58.3993 & \textbf{90.6} & 88.1 & 74.5 & \textbf{22.9} & 12.7 & 12.1 \\
500 & \textbf{51.4759} & 57.5518 & 63.4061 & \textbf{84.7} & 79.2 & 66.6 & \textbf{19.5} & 11.4 & 6.8
    & 500 & \textbf{46.3913} & 51.0218 & 57.2895 & \textbf{92.1} & 90.6 & 76.0 & \textbf{26.1} & 16.2 & 13.0 \\
\bottomrule
\end{tabular}
}
\end{table*}

\textbf{Randomness and Stability Across Patch Selection} Our patch-based refinement uses random patch locations and sizes, introducing stochasticity. To quantify this effect, we executed five independent runs per model. For each run and query budget, we computed the average $\ell_2$ distance across all pairs; we then summarized the five runs by reporting mean~$\pm$~standard deviation (std) of the average $\ell_2$ distance across all pairs. As seen in Table \ref{tab:rand_all_models_l2}, across all models and budgets, variability is small relative to the mean. The standard deviation is typically below $0.3$ (well under $1\%$ of the mean).

\begin{table*}[t]
\centering
\caption{Randomness analysis across five runs: $\ell_2$ (mean~$\pm$~std).}
\label{tab:rand_all_models_l2}
\begin{tabular}{rcccc}
\toprule
Budget & ResNet50 & ResNet101 & VGG16 & ViT \\
\midrule
50  & 88.1415~$\pm$~0.0254 & 87.0992~$\pm$~0.0637 & 90.6675~$\pm$~0.0553 & 78.7152~$\pm$~0.0277 \\
100 & 75.2155~$\pm$~0.2338 & 74.5036~$\pm$~0.0676 & 77.2608~$\pm$~0.1085 & 67.5654~$\pm$~0.0652 \\
150 & 67.3196~$\pm$~0.2301 & 66.6368~$\pm$~0.1022 & 69.1007~$\pm$~0.1479 & 60.6342~$\pm$~0.1187 \\
200 & 62.0338~$\pm$~0.2607 & 61.3794~$\pm$~0.1318 & 63.6530~$\pm$~0.1896 & 56.1255~$\pm$~0.0829 \\
250 & 58.2941~$\pm$~0.2458 & 57.6515~$\pm$~0.1356 & 59.7896~$\pm$~0.2565 & 53.0414~$\pm$~0.0537 \\
300 & 55.5589~$\pm$~0.1995 & 54.9321~$\pm$~0.1397 & 56.9954~$\pm$~0.2654 & 50.8059~$\pm$~0.0627 \\
400 & 51.9530~$\pm$~0.1907 & 51.3780~$\pm$~0.1564 & 53.4018~$\pm$~0.2513 & 47.8666~$\pm$~0.0654 \\
500 & 49.8352~$\pm$~0.1522 & 49.3863~$\pm$~0.1306 & 51.3763~$\pm$~0.2361 & 46.2528~$\pm$~0.0899 \\
\bottomrule
\end{tabular}
\end{table*}

\textbf{Lower Resolution Performance} 
We evaluate TEA, CGBA, CGBA-H and AHA at two lower input sizes, $128 \times 128$ and $64 \times 64$. The Tables \ref{tab:median_grouped_128} and \ref{tab:median_grouped_64} report the median $\ell_2$ distance to the source image achieved at fixed query budgets; lower is better. VGG-16 and ViT required standard architectural adjustments at reduced dimensions.

\begin{table*}[!ht]
\caption{Median $\ell_2$ distances at 128$\times$128.}
\label{tab:median_grouped_128}
\centering
\begin{tabular}{c c c c c c c c c c}
\toprule
\multirow{2}{*}{Query}
  & \multicolumn{4}{c}{ResNet50}
  & \multirow{2}{*}{Query}
  & \multicolumn{4}{c}{ResNet101} \\
\cmidrule(lr){2-5} \cmidrule(lr){7-10}
  & AHA & CGBA & CGBA-H & TEA
  &     & AHA & CGBA & CGBA-H & TEA \\
\midrule
50  & \textbf{52.812610} & 56.892204 & 56.892204 & 53.891792 & 50  & \textbf{52.325765} & 55.829930 & 55.830184 & 52.836270 \\
100 & 49.792591 & 52.638543 & 52.610928 & \textbf{44.656296} & 100 & 49.256166 & 51.690833 & 52.297024 & \textbf{43.908848} \\
150 & 47.580821 & 48.751302 & 48.900604 & \textbf{40.200085} & 150 & 47.165379 & 47.665892 & 48.747030 & \textbf{39.356266} \\
200 & 45.816245 & 47.670539 & 47.708520 & \textbf{37.588080} & 200 & 45.418667 & 46.499661 & 48.731472 & \textbf{36.744068} \\
250 & 44.242092 & 47.291906 & 47.416906 & \textbf{35.637354} & 250 & 43.904684 & 45.875023 & 48.286582 & \textbf{34.803333} \\
300 & 42.705044 & 45.310037 & 45.394565 & \textbf{34.083014} & 300 & 43.015997 & 43.947135 & 46.251110 & \textbf{33.364790} \\
350 & 41.704023 & 45.917060 & 46.053402 & \textbf{32.915845} & 350 & 42.285534 & 45.637712 & 48.112825 & \textbf{32.363468} \\
400 & 40.606535 & 43.876574 & 44.083020 & \textbf{31.711626} & 400 & 41.279571 & 43.208419 & 47.001603 & \textbf{31.200894} \\
\midrule
\multirow{2}{*}{Query}
  & \multicolumn{4}{c}{VGG16}
  & \multirow{2}{*}{Query}
  & \multicolumn{4}{c}{ViT} \\
\cmidrule(lr){2-5} \cmidrule(lr){7-10}
  & AHA & CGBA & CGBA-H & TEA
  &     & AHA & CGBA & CGBA-H & TEA \\
\midrule
50  & \textbf{53.742676} & 57.595045 & 57.595044 & 54.653168 & 50  & \textbf{49.717520} & 61.999604 & 61.999605 & 51.518260 \\
100 & 50.687319 & 53.132007 & 53.124912 & \textbf{45.101308} & 100 & 46.560914 & 56.845693 & 56.760291 & \textbf{42.468795} \\
150 & 48.452131 & 48.769476 & 48.688020 & \textbf{40.641963} & 150 & 44.133746 & 51.938031 & 51.849316 & \textbf{37.765469} \\
200 & 46.376375 & 47.648656 & 47.548296 & \textbf{37.729822} & 200 & 42.135314 & 52.195369 & 52.190025 & \textbf{34.882641} \\
250 & 44.719990 & 46.665404 & 46.692002 & \textbf{35.775914} & 250 & 40.692779 & 50.410471 & 50.458582 & \textbf{32.847132} \\
300 & 43.824315 & 44.433300 & 44.557708 & \textbf{34.574623} & 300 & 39.469095 & 47.456320 & 47.224463 & \textbf{31.225525} \\
350 & 42.589694 & 45.370771 & 45.405220 & \textbf{33.251792} & 350 & 38.537397 & 49.233241 & 49.101166 & \textbf{30.127403} \\
400 & 41.597263 & 43.446274 & 43.461510 & \textbf{32.151386} & 400 & 37.460420 & 47.532119 & 47.482755 & \textbf{29.117229} \\
\bottomrule
\end{tabular}
\end{table*}

\begin{table*}[!ht]
\caption{Median $\ell_2$ distances at 64$\times$64.}
\label{tab:median_grouped_64}
\centering
\begin{tabular}{c c c c c c c c c c}
\toprule
\multirow{2}{*}{Query}
  & \multicolumn{4}{c}{ResNet50}
  & \multirow{2}{*}{Query}
  & \multicolumn{4}{c}{ResNet101} \\
\cmidrule(lr){2-5} \cmidrule(lr){7-10}
  & AHA & CGBA & CGBA-H & TEA
  &     & AHA & CGBA & CGBA-H & TEA \\
\midrule
50  & \textbf{28.562640} & 31.263057 & 31.263184 & 29.709433 & 50  & \textbf{28.527201} & 31.240249 & 31.240327 & 29.828940 \\
100 & 26.583674 & 28.666067 & 28.734703 & \textbf{25.884418} & 100 & 26.545651 & 28.491136 & 28.543173 & \textbf{26.137666} \\
150 & 25.094041 & 26.041276 & 26.107465 & \textbf{23.768843} & 150 & 25.101155 & 25.824065 & 25.971276 & \textbf{23.934921} \\
200 & 23.449901 & 24.898503 & 25.231694 & \textbf{21.971898} & 200 & 23.850520 & 24.798062 & 25.058821 & \textbf{22.132178} \\
250 & 22.191710 & 24.117243 & 24.478960 & \textbf{20.671545} & 250 & 22.805084 & 23.948980 & 24.090509 & \textbf{20.863024} \\
\midrule
\multirow{2}{*}{Query}
  & \multicolumn{4}{c}{VGG16}
  & \multirow{2}{*}{Query}
  & \multicolumn{4}{c}{ViT} \\
\cmidrule(lr){2-5} \cmidrule(lr){7-10}
  & AHA & CGBA & CGBA-H & TEA
  &     & AHA & CGBA & CGBA-H & TEA \\
\midrule
50  & \textbf{28.318550} & 30.844857 & 30.837613 & 29.530338 & 50  & \textbf{30.081730} & 32.758422 & 32.758422 & 31.046581 \\
100 & 26.390916 & 28.302722 & 28.346767 & \textbf{25.606639} & 100 & 27.818672 & 29.806229 & 29.807765 & \textbf{27.077744} \\
150 & 24.731502 & 25.578313 & 25.651390 & \textbf{23.405260} & 150 & 26.072833 & 26.316862 & 26.340219 & \textbf{24.609781} \\
200 & 23.409042 & 24.719593 & 24.810609 & \textbf{21.847981} & 200 & 24.510567 & 24.761248 & 24.786895 & \textbf{22.262164} \\
250 & 22.453166 & 23.539063 & 23.687779 & \textbf{20.527478} & 250 & 23.248782 & 23.717796 & 23.829428 & \textbf{20.461638} \\
\bottomrule
\end{tabular}

\end{table*}

As observable in the results, at lower input resolution sizes, each patch modification distorts a larger fraction of the target image and thus contributes more to the overall $\ell_2$-norm reduction. Consequently, maintaining the adversariality of the image becomes much harder, and TEA therefore terminates earlier than for higher resolutions.

\textbf{Perceptual Metrics} While adversarial examples are generated to trick ML models, it is worth considering their impact on human perception \cite{carlini2017towards, fu2022autoda}. We report SSIM \cite{wang2004image} and FSIM \cite{zhang2011fsim} alongside distortion at a fixed budget of 400 queries. As shown in Table \ref{tab:fsim_ssim_400}, TEA achieves high distortion while slightly improving FSIM. A small drop in SSIM is expected, since TEA's perturbations are more structured. Prior image quality assessment (IQA) studies \cite{zhang2011fsim, fezza2019perceptual} indicate FSIM correlates more strongly with human perception than SSIM, which helps contextualise these differences.

\begin{table*}[t]
\caption{SSIM and FSIM at 400 queries on 1000 source---target pairs. \\Lower $\ell_2$ indicates smaller perturbations; higher SSIM/FSIM indicates greater perceptual similarity to the source image.}
\label{tab:fsim_ssim_400}
\centering
\begin{tabular}{c c c c c c c c c c}
\toprule
& \multicolumn{4}{c}{ResNet50} && \multicolumn{4}{c}{ResNet101} \\
\cmidrule(lr){2-5}\cmidrule(lr){7-10}
Metric & CGBA & CGBA-H & AHA & TEA && CGBA & CGBA-H & AHA & TEA \\
\midrule
$\ell_2$ & 85.579 & 71.479 & 64.172 & \textbf{51.9530} && 82.483 & 68.803 & 67.051 & \textbf{51.3780} \\
SSIM & \textbf{0.531765} & 0.530459 & 0.231784 & 0.478806 && 0.525847 & \textbf{0.528782} & 0.223511 & 0.491950 \\
FSIM & 0.251337 & 0.252585 & 0.232117 & \textbf{0.294672} && 0.252746 & 0.251235 & 0.223831 & \textbf{0.288153} \\
\midrule
& \multicolumn{4}{c}{VGG16} && \multicolumn{4}{c}{ViT} \\
\cmidrule(lr){2-5}\cmidrule(lr){7-10}
Metric & CGBA & CGBA-H & AHA & TEA && CGBA & CGBA-H & AHA & TEA \\
\midrule
$\ell_2$ & 89.880 & 70.559 & 68.538 & \textbf{53.4018} && 72.839 & 61.300 & 59.857 & \textbf{47.8666} \\
SSIM & \textbf{0.534783} & 0.532189 & 0.231784 & 0.457879 && 0.580527 & \textbf{0.582544} & 0.201767 & 0.548110 \\
FSIM & 0.251129 & 0.251239 & 0.232117 & \textbf{0.305919} && 0.220927 & 0.220213 & 0.202171 & \textbf{0.257571} \\
\bottomrule
\end{tabular}
\end{table*}

\begin{figure*}[t]
    \centering
    \begin{minipage}[b]{0.24\textwidth}
        \centering
        \includegraphics[width=\textwidth]{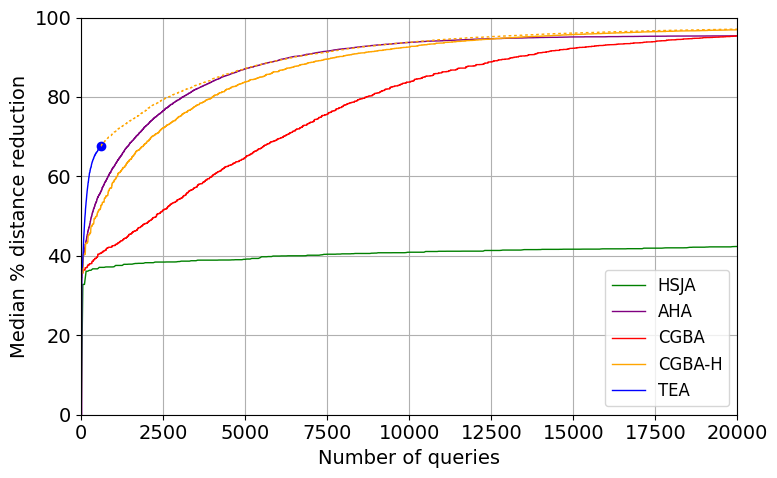}
        \par\small ResNet50
    \end{minipage}\hfill
    \begin{minipage}[b]{0.24\textwidth}
        \centering
        \includegraphics[width=\textwidth]{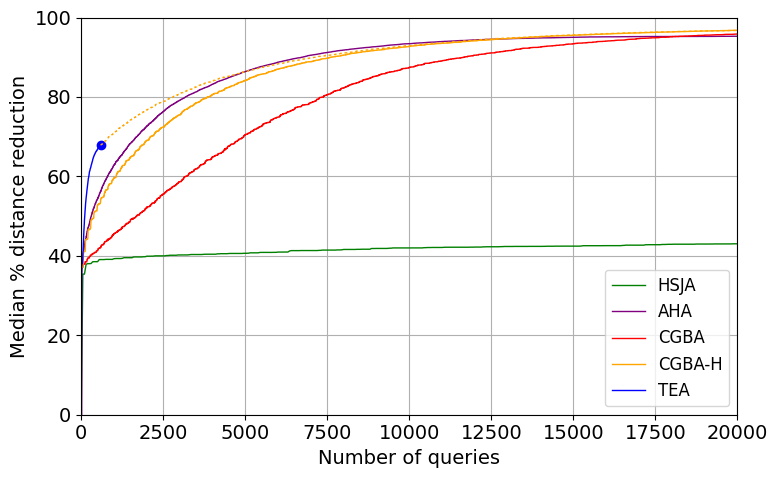}
        \par\small ResNet101
    \end{minipage}\hfill
    \begin{minipage}[b]{0.24\textwidth}
        \centering
        \includegraphics[width=\textwidth]{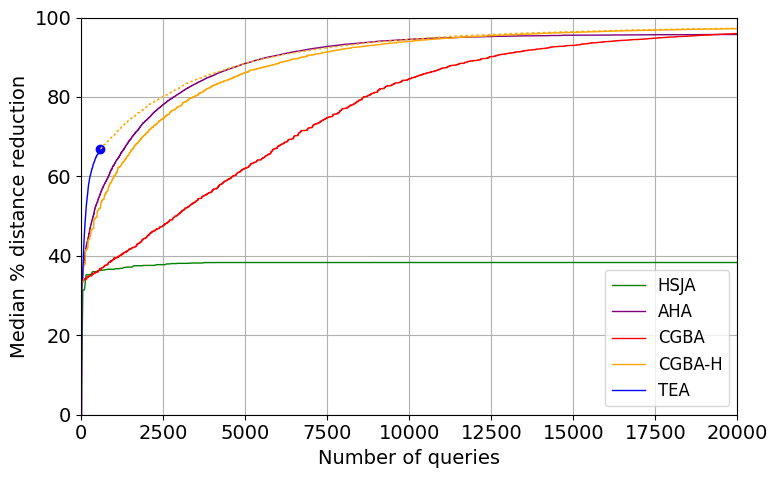}
        \par\small VGG16
    \end{minipage}\hfill
    \begin{minipage}[b]{0.24\textwidth}
        \centering
        \includegraphics[width=\textwidth]{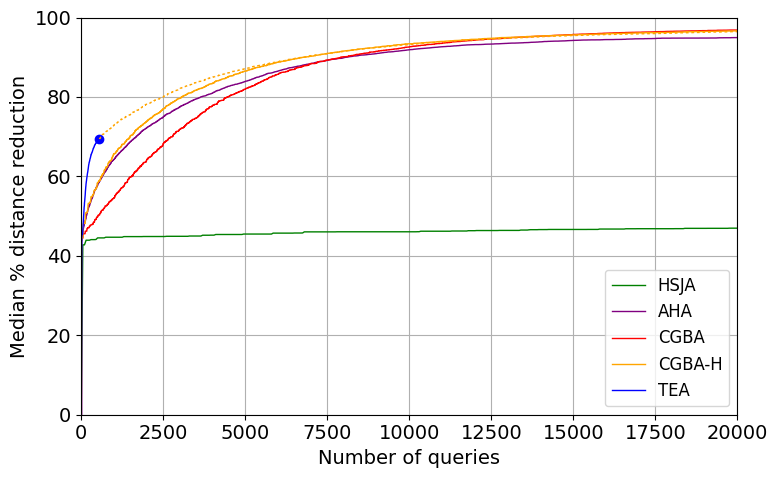}
        \par\small ViT
    \end{minipage}
\caption{Comparison of median percentage decrease in $\ell_2$ distance across different architectures --- higher values indicate a more effective reduction method.}
    \label{fig:median_grouped_20000} 
\end{figure*}

\textbf{High‐Query Performance with CGBA-H refinement.} Our hybrid TEA+CGBA‑H strategy consistently matches or surpasses the state-of-the-art. Table \ref{tab:median_grouped_20000} lists the median $\ell_2$ distances upto 20000 queries.  Note that since CGBA-H is randomness-dependent, it doesn't re-explore when trapped in a narrow decision space and its stability is subject to local decision boundary geometry. Figure \ref{fig:median_grouped_20000} illustrates the median percentage reduction in the distance between the images.

\begin{table*}[t]
  \caption{Median $\ell_2$ distances across different architectures. \\Here, TEA$^{\ast}$ indicates using TEA until the turning point, and refining further with CGBA-H.}
  \label{tab:median_grouped_20000}
  \centering
    \begin{tabular}{c c c c c c c c c c c c}
      \toprule
      \multirow{2}{*}{Query} 
        & \multicolumn{5}{c}{ResNet50} 
        & \multirow{2}{*}{Query} 
        & \multicolumn{5}{c}{ResNet101} \\
      \cmidrule(lr){2-6} \cmidrule(lr){8-12}
        & HSJA   & CGBA   & CGBA-H & AHA    & TEA$^{\ast}$ 
        &       & HSJA   & CGBA   & CGBA-H & AHA    & TEA$^{\ast}$ \\
      \midrule
1000  & 86.800 & 80.709 & 58.815 & 52.069  & \textbf{41.108} & 1000  & 86.162 & 76.005  & 55.824 & 51.642 & \textbf{40.090} \\
2500  & 85.262 &69.031& 39.701  & 33.165 & \textbf{29.747} & 2500  & 85.314 & 63.477 &  38.043 & 33.259& \textbf{29.965} \\
5000  & 83.345 & 48.703  &22.835 & 18.394 & \textbf{18.338} & 5000  & 84.202 &  41.437  &22.470 & \textbf{18.780} &18.932 \\
10000 & 81.623 & 22.094  & 10.214 &  \textbf{8.597} & 8.798  & 10000 & 82.443 & 17.623 &  10.341 & \textbf{9.021} & 9.864 \\
15000 & 80.642 & 10.999 &   5.908& 6.757  & \textbf{5.429} & 15000 & 81.873 & 8.936  & 6.341 & 6.937 & \textbf{6.281} \\
20000 & 79.605 &  6.594 & 4.185  & 6.449   & \textbf{4.011} & 20000 & 80.886 &  5.689  &\textbf{4.507}  &  6.644 & 4.560 \\
\midrule
      \multirow{2}{*}{Query} 
        & \multicolumn{5}{c}{VGG16} 
        & \multirow{2}{*}{Query} 
        & \multicolumn{5}{c}{ViT} \\
      \cmidrule(lr){2-6} \cmidrule(lr){8-12}
        & HSJA    & CGBA   & CGBA-H & AHA    & TEA$^{\ast}$ 
        &       & HSJA   & CGBA   & CGBA-H & AHA    & TEA$^{\ast}$ \\
      \midrule
1000  & 87.451 & 85.772 & 56.388  &  52.361& \textbf{41.578} & 1000  & 79.212& 64.516& 49.394  & 49.541 & \textbf{38.482} \\
2500  & 86.346 & 74.963 & 35.686  & 30.429 & \textbf{27.984} & 2500  & 78.739  &   44.789& 32.573& 34.762 & \textbf{28.134} \\
5000  & 85.219 & 53.869 &  19.753 & 16.076 & \textbf{16.050} & 5000  & 78.170 & 25.118 & 18.566& 22.197  & \textbf{18.507} \\
10000 & 85.219 &20.792 &   8.558 &\textbf{7.456} & 7.626 & 10000 & 77.786 & 10.319 & \textbf{9.299} & 11.267 & 9.416 \\
15000 & 85.219 & 9.455  & 5.306 & 6.238  & \textbf{4.979} & 15000 & 77.087 & \textbf{5.920}& 5.987  &  8.180  & 6.301 \\
20000 & 85.219 & 5.687  &  3.985 & 6.042 & \textbf{3.823} & 20000 & 76.727 & \textbf{4.258} & 4.635  &  7.404 & 4.955 \\
\bottomrule
\end{tabular}
\end{table*}

\begin{figure*}[t]
    \centering
    \begin{minipage}[b]{0.24\textwidth}
        \centering
        \includegraphics[width=\textwidth]{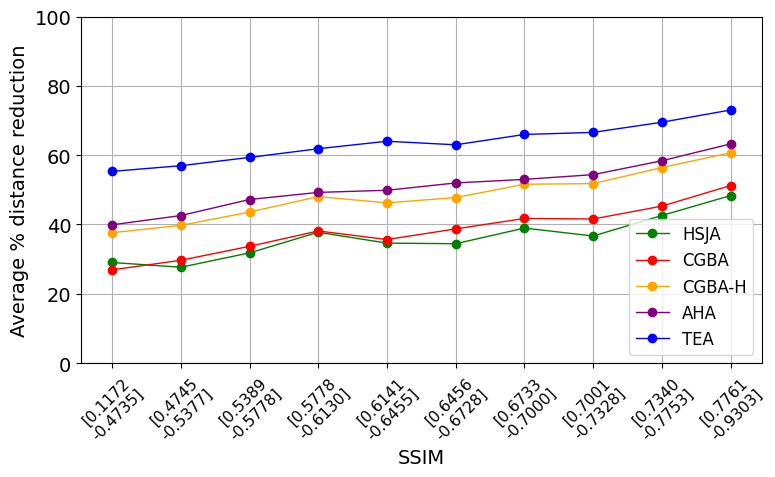}
        \par\small ResNet50
    \end{minipage}\hfill
    \begin{minipage}[b]{0.24\textwidth}
        \centering
        \includegraphics[width=\textwidth]{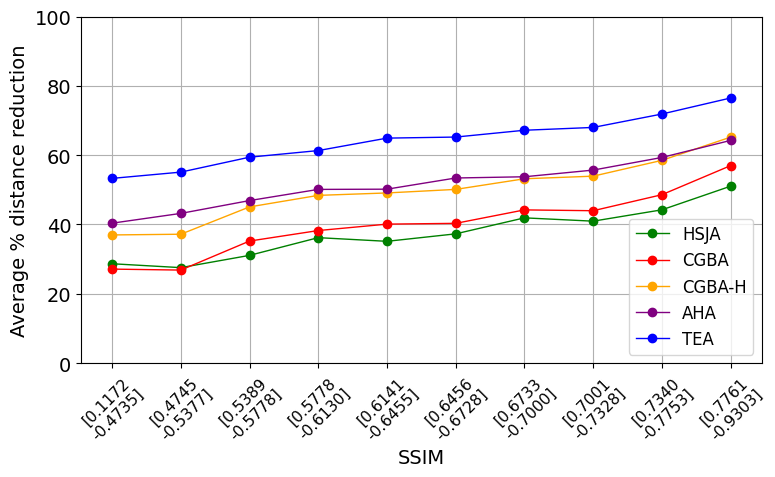}
        \par\small ResNet101
    \end{minipage}\hfill
    \begin{minipage}[b]{0.24\textwidth}
        \centering
        \includegraphics[width=\textwidth]{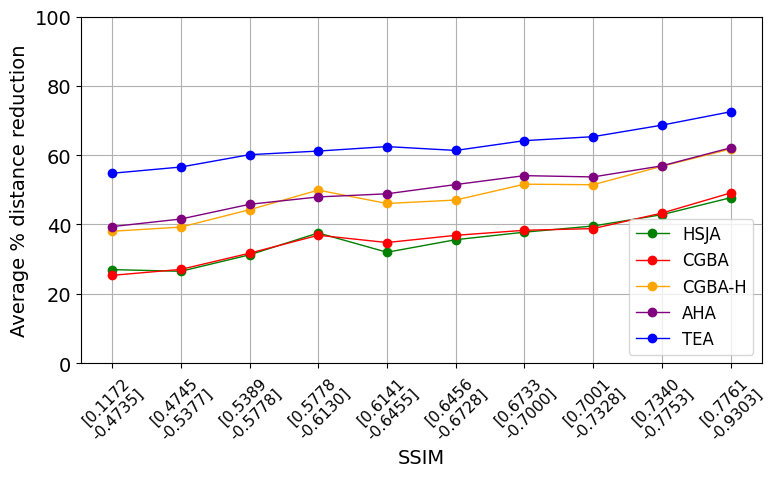}
        \par\small VGG16
    \end{minipage}\hfill
    \begin{minipage}[b]{0.24\textwidth}
        \centering
        \includegraphics[width=\textwidth]{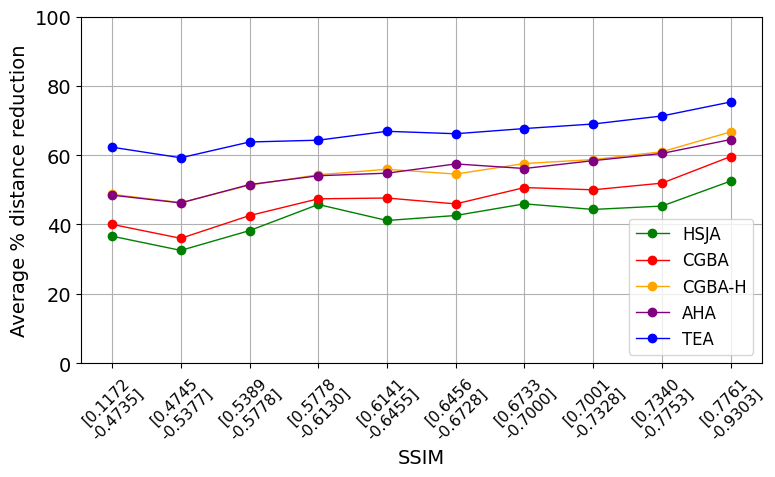}
        \par\small ViT
    \end{minipage}
\caption{Effect of source–target structural similarity on attack performance. We group image pairs into ten bins by SSIM and report the average percentage decrease in $\ell_2$ distance after 400 queries. All methods benefit similarly from higher structural similarity.}
    \label{fig:ssim} 
\end{figure*}

\textbf{Source-Target Similarity Analysis.}
To examine how the performance of TEA depends on the similarity between source and target images, we group the $1000$ source-image pairs by structural similarity. For each pair $(x_s, x_t)$ we compute the SSIM score and sort all pairs by SSIM, partitioning them into ten equally sized bins. Within each bin, and model, we measure the percentage $\ell_2$ reduction at a fixed budget of $400$ queries. We then average this quantity over all pairs in the same bin and plot the resulting curves (Figure~\ref{fig:ssim}). We see that all methods have relatively identical performance variability, with improved performance across structurally similar source-target image pairs. This indicates that structural similarity does not favor TEA over the other methods.  

\begin{figure*}[t]
    \centering
    \begin{minipage}[b]{0.24\textwidth}
        \centering
        \includegraphics[width=\textwidth]{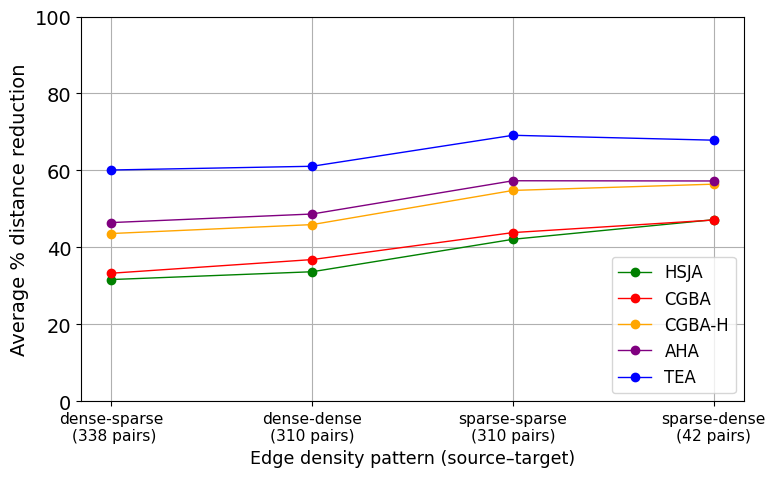}
        \par\small ResNet50
    \end{minipage}\hfill
    \begin{minipage}[b]{0.24\textwidth}
        \centering
        \includegraphics[width=\textwidth]{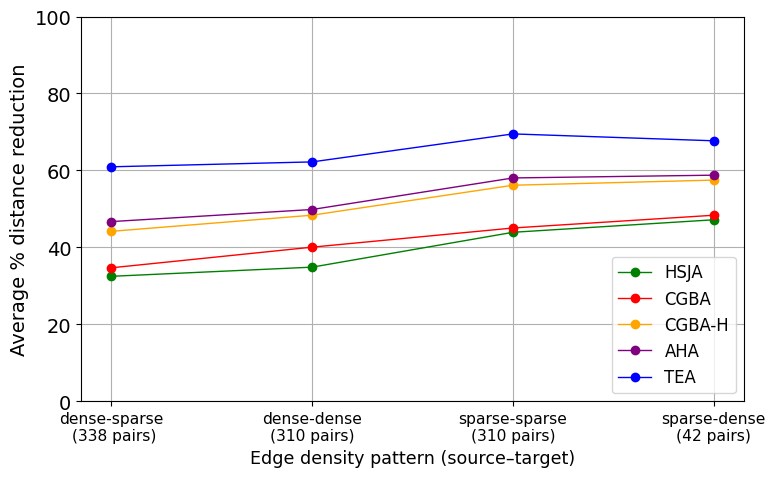}
        \par\small ResNet101
    \end{minipage}\hfill
    \begin{minipage}[b]{0.24\textwidth}
        \centering
        \includegraphics[width=\textwidth]{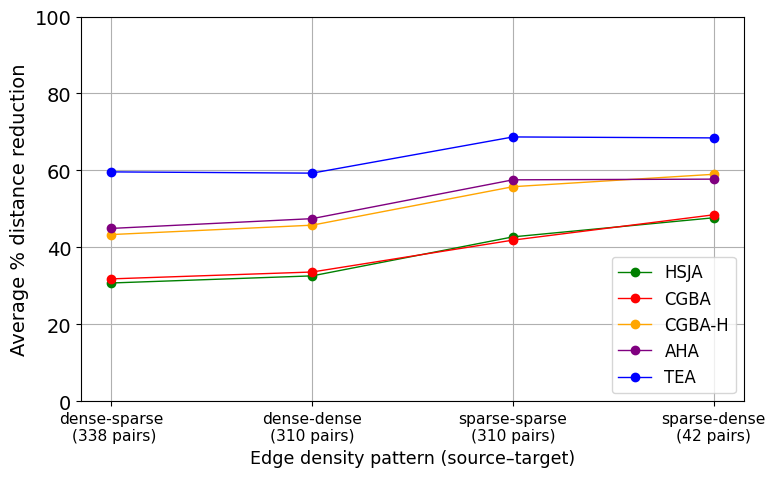}
        \par\small VGG16
    \end{minipage}\hfill
    \begin{minipage}[b]{0.24\textwidth}
        \centering
        \includegraphics[width=\textwidth]{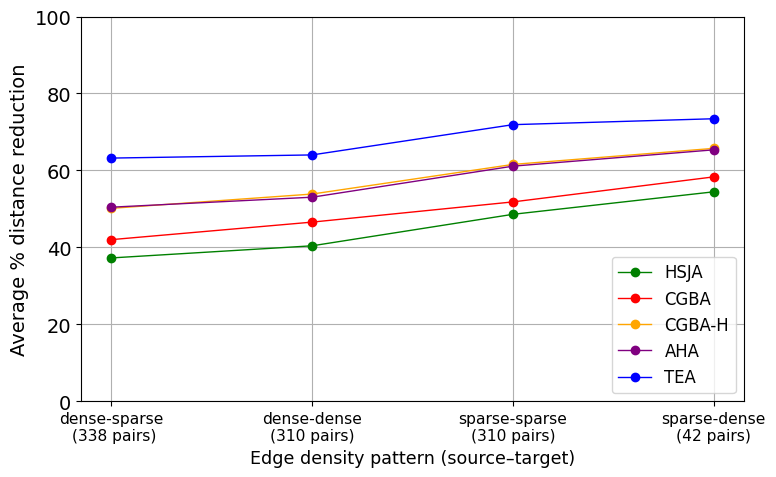}
        \par\small ViT
    \end{minipage}
\caption{Effect of source–target edge density on attack performance. Each image is classified as \emph{dense} or \emph{sparse} based on a Sobel edge–density threshold, inducing four edge–pattern regimes (dense–dense, dense–sparse, sparse–dense, sparse–sparse). For each regime, and model, we report the average percentage decrease in $\ell_2$ distance after 400 queries, illustrating how performance varies with the edge richness of the source–target pairs.}
    \label{fig:density} 
\end{figure*}

\textbf{Edge–Density Analysis.}
To assess how TEA behaves under different levels of structural detail in the source and target images, we perform an edge–density based stratification of the source–target pairs. For each image, we convert it to grayscale, compute Sobel gradients, and form the normalized gradient magnitude $\tilde{g}(p) \in [0,1]$ for each pixel $p$. Pixels with $\tilde{g}(p) > 0.2$ are treated as edge pixels, and the edge density $d(x)$ is defined as the fraction of pixels classified as edges. We collect $d(x_s)$ and $d(x_t)$ for all source-target image pairs and set a global dense/sparse threshold $\tau_{\text{dense}}$ to be the median density over all $2000$ images, yielding $\tau_{\text{dense}} \approx 0.110242$ in our case. Images with $d(x) > \tau_{\text{dense}}$ are labeled \emph{dense}, and the rest \emph{sparse}. Each pair $(x_s, x_t)$ is thus assigned to one of four edge–pattern regimes: dense–sparse, dense–dense, sparse–sparse, or sparse–dense. For each regime, and model, we report the average percentage $\ell_2$ reduction (in \%) at a fixed budget of $400$ queries. The resulting curves in Figure~\ref{fig:density} show that performance variability is similar across methods, with higher success when moving towards a sparse source image than a dense one, and that TEA maintains a consistent advantage across all regimes. 

\textbf{Zero-Shot CLIP.}
To assess whether TEA also transfers to more modern vision-language models, we additionally evaluate it against a zero-shot CLIP classifier on the ImageNet validation set. We use the ViT-B/32 variant of CLIP \cite{radford2021learning,dosovitskiy2020image} in its standard zero-shot configuration. Following the usual protocol, we construct a linear zero-shot head by encoding multiple natural-language templates for each ImageNet class (e.g., ``a photo of a \{\}.'', ``a close-up photo of a \{\}.'') with the CLIP text encoder, normalizing and averaging the resulting embeddings to obtain a single prototype per class. Stacking these prototypes yields a fixed weight matrix, and logits are obtained by taking scaled dot products between normalized image features and these class prototypes. Our implementation relies on the official CLIP PyTorch code-base and its recommended preprocessing.

\begin{figure}[t]
\centering
\resizebox{\columnwidth}{!}{%
\begin{tabular}{@{}cc@{}}
\includegraphics[width=.48\linewidth]{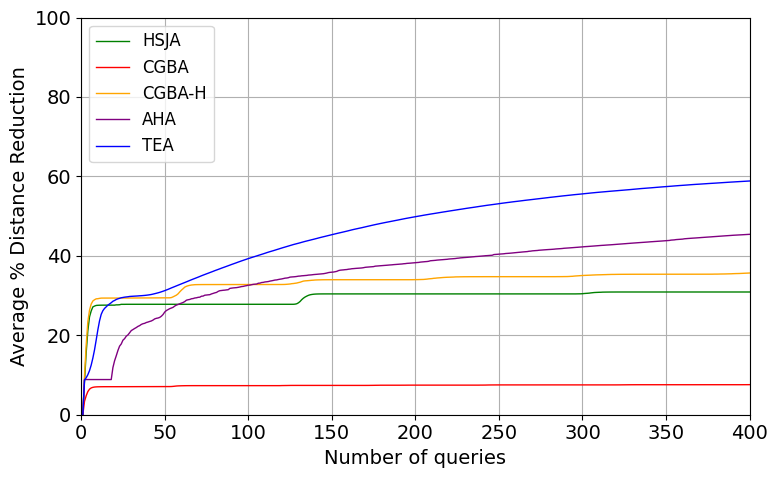} &
\includegraphics[width=.48\linewidth]{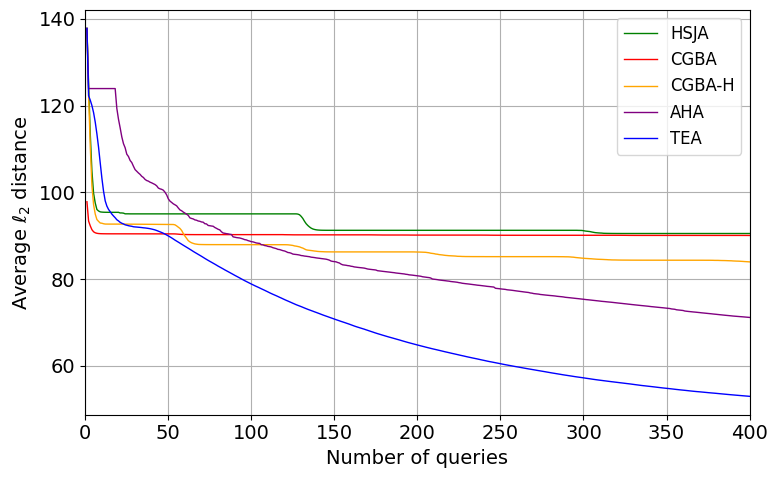}
\end{tabular}%
}
\caption{Comparison on a zero-shot CLIP (ViT-B/32) classifier on ImageNet. Left: average percentage decrease in $\ell_2$ distance against number of queries, with higher values indicating a more effective reduction method. Right: average $\ell_2$ distance against number of queries, with lower values indicating a more effective reduction method.}

\label{fig:clip}
\end{figure}

In Figure~\ref{fig:clip}, we show the average percentage decrease in $\ell_2$ distance and the average $\ell_2$ distance as a function of the query budget. TEA retains a clear advantage, consistently achieving larger perturbation reductions than the baselines within the same limited query budget, indicating that its strategy remains effective even for modern zero-shot vision-language models.

\begin{figure*}[t]
    \centering
    \begin{minipage}[b]{0.24\textwidth}
        \centering
        \includegraphics[width=\textwidth]{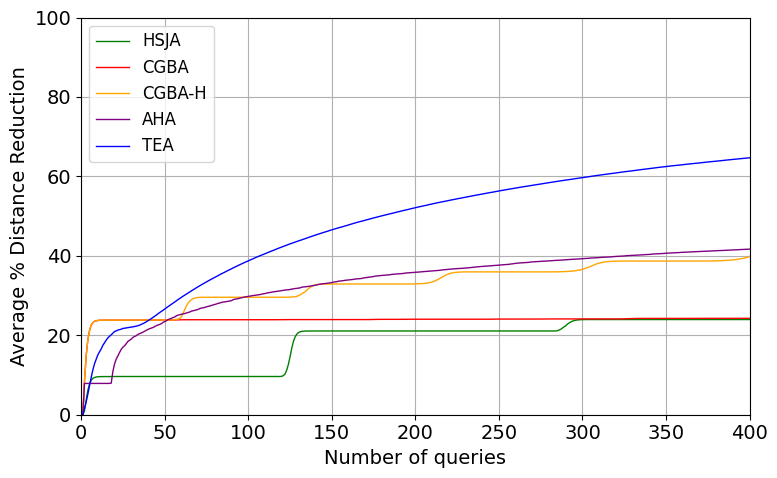}
        \par\small ResNet50
    \end{minipage}\hfill
    \begin{minipage}[b]{0.24\textwidth}
        \centering
        \includegraphics[width=\textwidth]{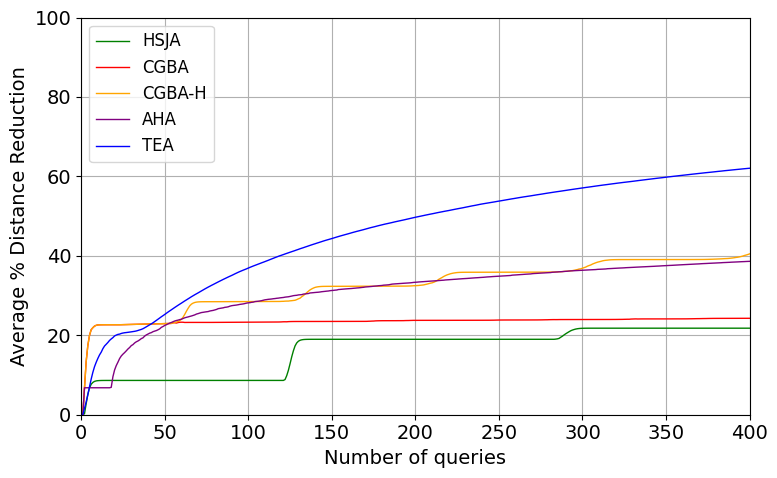}
        \par\small ResNet101
    \end{minipage}\hfill
    \begin{minipage}[b]{0.24\textwidth}
        \centering
        \includegraphics[width=\textwidth]{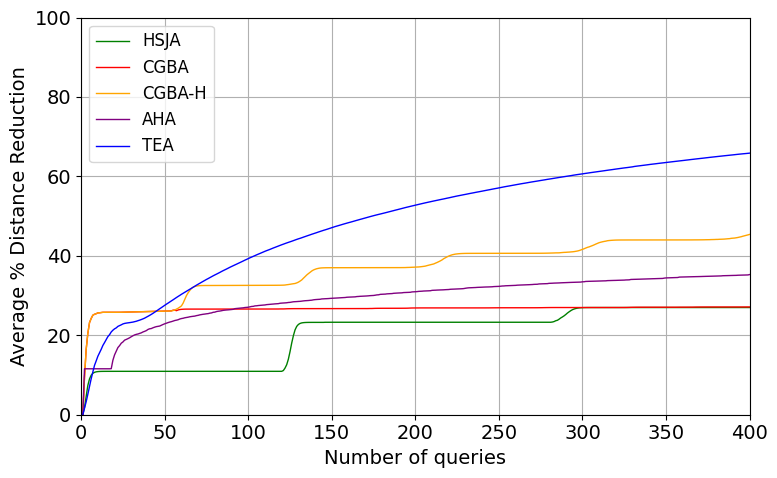}
        \par\small VGG16
    \end{minipage}\hfill
    \begin{minipage}[b]{0.24\textwidth}
        \centering
        \includegraphics[width=\textwidth]{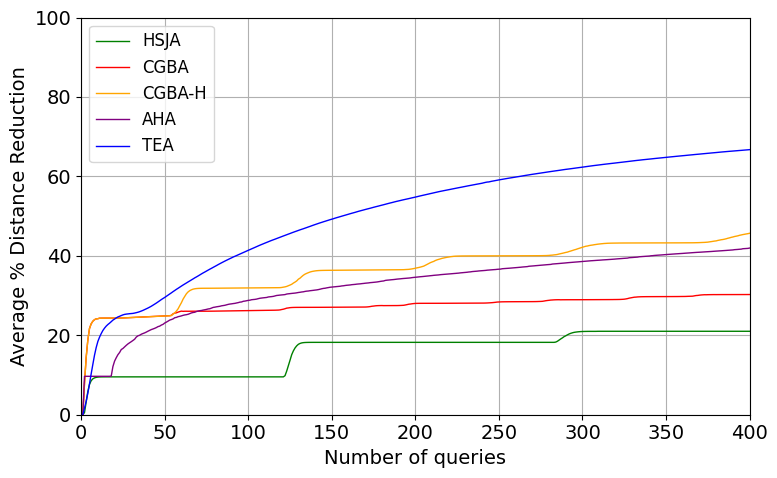}
        \par\small ViT
    \end{minipage}
    \caption{Average $\ell_2$ distance reduction across different architectures in a low-query regime on the CIFAR-100 dataset.}
    \label{fig:cifar} 
\end{figure*}

\begin{figure*}[t]
    \centering
    \begin{minipage}[b]{0.24\textwidth}
        \centering
        \includegraphics[width=\textwidth]{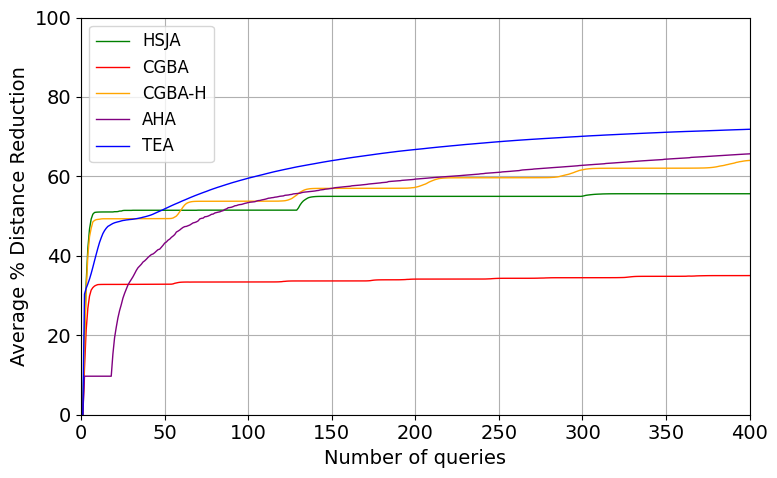}
        \par\small ResNet50
    \end{minipage}\hfill
    \begin{minipage}[b]{0.24\textwidth}
        \centering
        \includegraphics[width=\textwidth]{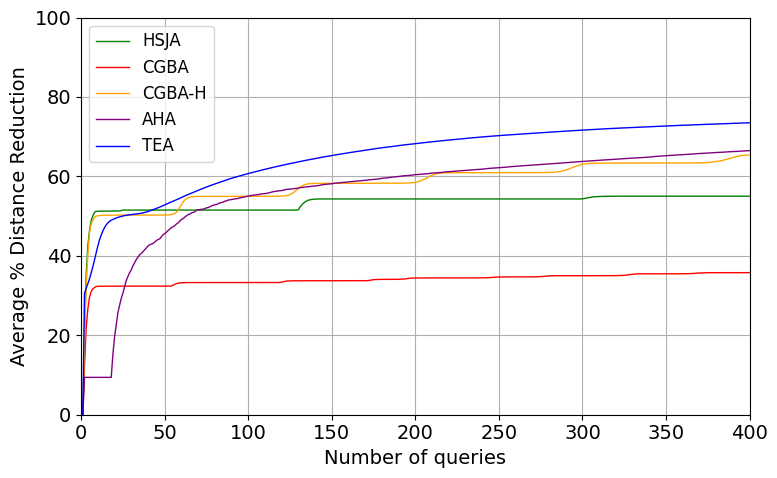}
        \par\small ResNet101
    \end{minipage}\hfill
    \begin{minipage}[b]{0.24\textwidth}
        \centering
        \includegraphics[width=\textwidth]{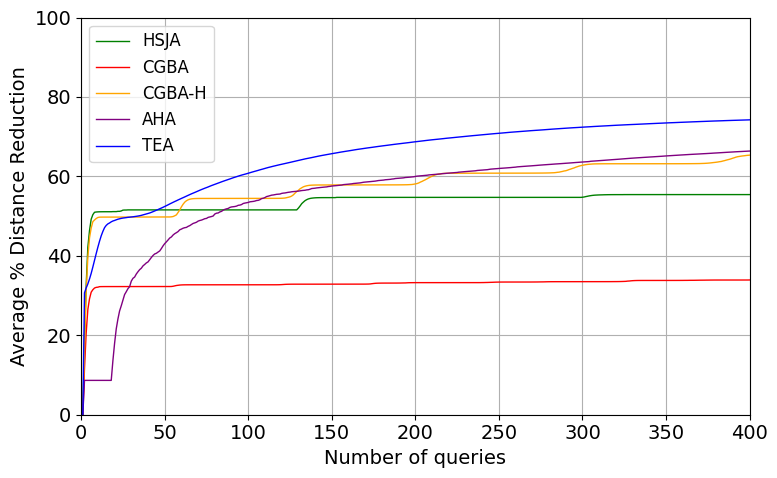}
        \par\small VGG16
    \end{minipage}\hfill
    \begin{minipage}[b]{0.24\textwidth}
        \centering
        \includegraphics[width=\textwidth]{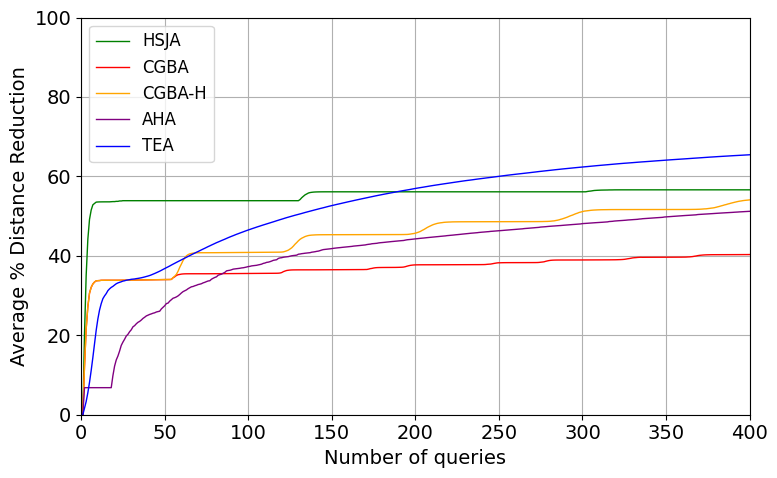}
        \par\small ViT
    \end{minipage}
    \caption{Average $\ell_2$ distance reduction across different architectures in a low-query regime on the Intel Image Classification dataset.}
    \label{fig:intel} 
\end{figure*}

\textbf{Other Datasets.}
To further assess the robustness and generality of TEA beyond ImageNet, we additionally evaluate all tested methods on two standard classification benchmark datasets: CIFAR-100 \cite{krizhevsky2009learning} and the Intel Image Classification dataset \cite{intel_scene}. CIFAR-100 contains 100 object categories with lower-resolution images while the Intel dataset consists of six scene categories (e.g., buildings, forest, sea). In both cases, we again attack the four models, ResNet-50, ResNet-101, VGG-16, and ViT, and measure performance in terms of the relative $\ell_2$-distance reduction between the source and target images. Figures~\ref{fig:cifar} and \ref{fig:intel} report the average $\ell_2$-distance reduction at a fixed low-query budget of $400$ across all four models. Consistent with our results on the ImageNet dataset, TEA achieves larger perturbation reductions on both datasets, indicating that its edge-aware strategy remains effective across diverse data distributions and resolutions.

\section{Conclusion}\label{sec:discussion}
In this work, we introduced \emph{TEA}, a targeted, hard-label, black-box adversarial attack that leverages edge information from a target image to efficiently produce adversarial examples perceptually closer to a source image in low-query settings. TEA initially employs a global search that preserves prominent edge structures across the target image, subsequently refining the perturbations via patch-wise updates. Empirical results demonstrate that TEA significantly reduces the $\ell_2$ distance between adversarial and source images, requiring over \emph{70\% fewer queries} compared to current state-of-the-art methods. In addition, it also achieves reduced AUC and high ASR scores, indicating a consistently rapid reduction in distance to the source image across different models.

\textbf{Limitations.} TEA is specifically designed for the targeted, hard-label, low-query black-box setting, and its main advantage lies in rapidly improving performance under tight query budgets. When TEA is used merely as an initialization and followed by stronger high-query attacks in regimes where many queries are available, we observe that the final performance is comparable to that of the underlying state-of-the-art methods, rather than strictly better. In addition, while TEA preserves global edge structure and achieves low $\ell_2$ distances, the patch-wise updates can still introduce localized artifacts that are visible to human observers. This limitation is shared with other norm-bounded attacks in general, but in TEA the patch-based nature of the perturbations can make these local changes particularly noticeable. 

\textbf{Future Work.} Several extensions could enhance TEA: (a) substituting edge information with other features such as textures, color distributions, or high-frequency components; (b) training a surrogate model based on query data already collected to guide subsequent perturbations and (c) developing defense mechanisms capable of mitigating attacks based on structure-preserving perturbations.

\section*{Code Availability}
Our code is available at the following URL: \url{https://github.com/mdppml/TEA}.

\section*{Code of Ethics and Broader Impact Statement}
We evaluate TEA exclusively on the publicly available ImageNet-1K validation set, which is distributed for non-commercial research and educational use under the ImageNet access agreement. The dataset contains no personally identifiable information, and was used in accordance with the license terms. As TEA reduces the number of required queries for targeted hard-label attacks in a low-query setting, it could be leveraged to more rapidly craft adversarial inputs against commercial or safety-critical vision systems. We encourage practitioners to adopt defense mechanisms that go beyond detecting frequency-based perturbations, explicitly incorporating checks for edge-informed distortions, and defenses attuned to other structural features, to guard against similar attacks.

\section*{LLM Usage Considerations}
LLMs were used for editorial purposes in this manuscript, and all outputs were inspected by the authors to ensure accuracy and originality. All technical ideas, experiments, analyses, and citations were conceived, implemented, and validated by the authors.

\section*{Acknowledgments}

This research was supported by the German Federal Ministry of Education and Research (BMBF) under project number 01ZZ2010 and partially funded through grant 01ZZ2316D (PrivateAIM). The authors acknowledge the usage of the Training Center for Machine Learning (TCML) cluster at the University of Tübingen.

\FloatBarrier


\bibliographystyle{IEEEtran}
\bibliography{main}

\end{document}